\documentclass[sn-mathphys-num]{sn-jnl}% Math and Physical Sciences Numbered Reference Style 
\usepackage{graphicx}%
\usepackage{multirow}%
\usepackage{amsmath,amssymb,amsfonts}%
\usepackage{amsthm}%
\usepackage{mathrsfs}%
\usepackage[title]{appendix}%
\usepackage{xcolor}%
\usepackage{textcomp}%
\usepackage{manyfoot}%
\usepackage{booktabs}%
\usepackage{algorithm}%
\usepackage{algorithmicx}%
\usepackage{algpseudocode}%
\usepackage{listings}%
\usepackage[inline]{enumitem}
\usepackage{array}
\usepackage{subfig}
\usepackage{natbib}
\usepackage{lineno}
\usepackage{changes}

\newcommand\wordcount{
    \immediate\write18{texcount -sub=section \jobname.tex  | grep "Section" | sed -e 's/+.*//' | sed -n \thesection p > 'count.txt'}
(\input{count.txt}words)}

\theoremstyle{thmstyleone}%
\theoremstyle{thmstyletwo}%

\theoremstyle{thmstylethree}%

\newcommand{\eg}{e.\,g., }
\newcommand{\ie}{i.\,e., }
\providecommand{\tabularnewline}{\\}

\begin{document}

% \input{response}

% \title[Aligning generalisation Between Humans and Machines]{Aligning generalisation Between Humans and Machines}
\title[Aligning Generalisation Between Humans and Machines]{Aligning Generalisation Between Humans and Machines}

%TODO: add all persons who contributed as author names
%TODO: finalize author list and order
%

% Evaluation group contributors:
% Human-AI team contributors:
% Types group contributors: 

%Current order: Filip as main coordinator is first, then it is alphabetically
%Current authors: persons who coordinated the writing, either subchapters or final version, are mentioned, obviously incomplete as others contributed, this needs to be extended either by persons themselves or the chaper coordinators
%%=============================================================%%

\author*[1]{\fnm{Filip} \sur{Ilievski}}\email{f.ilievski@vu.nl}

\author[2]{\fnm{Barbara} \sur{Hammer}}

\author[1]{\fnm{Frank} \sur{van Harmelen}}

\author[2]{\fnm{Benjamin} \sur{Paassen}}

\author[3]{\fnm{Sascha} \sur{Saralajew}}

\author[4]{\fnm{Ute} \sur{Schmid}}

\author[5]{\fnm{Michael}\sur{Biehl}} 

\author[6]{\fnm{Marianna} \sur{Bolognesi}}

\author[7]{\fnm{Xin Luna} \sur{Dong}}

\author[3,8]{\fnm{Kiril} \sur{Gashteovski}}

\author[9]{\fnm{Pascal} \sur{Hitzler}}

\author[10]{\fnm{Giuseppe} \sur{Marra}}

\author[11,12]{\fnm{Pasquale} \sur{Minervini}}

\author[13]{\fnm{Martin} \sur{Mundt}}

\author[14]{\fnm{Axel-Cyrille} \sur{Ngonga Ngomo}}

\author[15]{\fnm{Alessandro} \sur{Oltramari}}

\author[16]{\fnm{Gabriella} \sur{Pasi}}

\author[17]{\fnm{Zeynep G.} \sur{Saribatur}}

\author[18]{\fnm{Luciano} \sur{Serafini}}

\author[19]
{\fnm{John} \sur{Shawe-Taylor}}

\author[20,21]{\fnm{Vered} \sur{Shwartz}}

\author[22]{\fnm{Gabriella} \sur{Skitalinskaya}}

\author[23]{\fnm{Clemens} \sur{Stachl}}

\author[10]{\fnm{Gido M.} \sur{van de Ven}}

\author[24,25]{\fnm{Thomas} \sur{Villmann}}

\affil[1]{Vrije Universiteit Amsterdam}

\affil[2]{University of Bielefeld}

\affil[3]{NEC Laboratories Europe}

\affil[4]{University of Bamberg}

\affil[5]{University of Groningen} %, The Netherlands}

\affil[6]{Università di Bologna}

\affil[7]{Meta Reality Labs}

\affil[8]{CAIR, Ss. Cyril and Methodius University of Skopje}

\affil[9]{Kansas State University}

\affil[10]{KU Leuven}

\affil[11]{University of Edinburgh}

\affil[12]{Miniml.AI}

\affil[13]{University of Bremen}

% \affil[13]{Technical University of Darmstadt}

% \affil[14]{The Hessian Center for Artificial Intelligence: hessian.AI}

\affil[14]{Paderborn University}

\affil[15]{Carnegie Bosch Institute}

\affil[16]{Università degli Studi di Milano Bicocca}

\affil[17]{TU Wien}

\affil[18]{Fondazione Bruno Kessler}

\affil[19]{University College London}

\affil[20]{University of British Columbia}

\affil[21]{Vector Institute}

\affil[22]{Duolingo}

\affil[23]{University of St. Gallen, Institute of Behavioral Science and Technology}

% \affil[24]{Ludwig-Maximilians-Universität München}

\affil[24]{University of Applied Sciences Mittweida}

\affil[25]{Technical University Freiberg}

\clearpage
\newpage
\abstract{Recent advances in AI---including generative approaches---have resulted in technology that can support humans in scientific discovery and forming decisions, but may also disrupt democracies and target individuals. The responsible use of AI and its participation in human-AI teams increasingly shows the need for \textit{AI alignment}, that is, to make AI systems act according to our preferences. A crucial yet often overlooked aspect of these interactions is the different ways in which humans and machines \textit{generalise}. In cognitive science, human generalisation commonly involves abstraction and concept learning. In contrast, AI generalisation encompasses out-of-domain generalisation in machine learning, rule-based reasoning in symbolic AI, and abstraction in neurosymbolic AI\@. In this perspective paper, we combine insights from AI and cognitive science to identify key commonalities and differences across three dimensions: notions of, methods for, and evaluation of generalisation. We map the different conceptualisations of generalisation in AI and cognitive science along these three dimensions and consider their role for alignment in human-AI teaming. This results in interdisciplinary challenges across AI and cognitive science that must be tackled to provide a foundation for effective and cognitively supported alignment in human-AI teaming scenarios.
\footnotemark[0]
\footnotetext[0]{A preprint of this manuscript is available on arXiv~\citep{preprint}.}
}

\keywords{generalisation, human-AI teaming, alignment}



\maketitle

\newpage
\tableofcontents
\newpage

% \linenumbers

\section{Introduction}

Recent technological developments in artificial intelligence (AI) have resulted in advances that allow for meaningful support of humans in complex tasks such as scientific discovery and decision making \citep{jumper2021highly}.
%, as exemplified by its recent use for predicting protein structures \citep{jumper2021highly}. 
In contrast, AI can also potentially disrupt democracies and target individuals \citep{ferrara2024genai}.
%\citep{ferrara2024genai}, as shown by the deepfake audio of President Biden in the New Hampshire primary. 
The responsible use of AI increasingly highlights the need for \textit{AI alignment}, which according to \citet{10.5555/3692070.3692292} aims to 
\emph{``make AI systems act according to our preferences''}.\footnote{Notably, alignment does not require the corresponding capabilities that enable those preferences to be realised to be similar between humans and AI.}
% without an expectation that the capabilities to achieve those are largely different~\cite{}. 
The alignment of humans and AI is essential for effective human-AI teaming, especially in complex scenarios where neither humans nor AI perform well on their own~\citep{metcalfe2021systemic}.
For example, AI can help invent novel biomedical application hypotheses following the objectives and guidance of a scientist~\citep{gottweis2025towards}. Alternatively, humans can iteratively edit samples provided by an AI model to improve its accuracy and trustworthiness in classifying skin cancer~\citep{donnelly2025rashomon}. These examples demonstrate how human-AI teams benefit from their complementary capabilities as they strive for similar goals. Besides human-AI teaming as a natural use case, aligning AI with humans is necessary for its safe use~\citep{bengio2025international} and demonstrable adherence to requirements for accountability, privacy, and transparency in legal frameworks such as the EU's AI Act~\citep{10.1145/3665322}.

% Humans edit iteratively the samples provided by a prototype-based model to improve the accuracy and trustworthiness; skin cancer classification.

% This enables humans to have a mental model of 
% Namely, while the methods of AI and humans may be vastly different, humans must have a mental model of 
% expectations of the AI that performs a task or fulfils a purpose, which is essential for an effective human-AI interaction. 
% The need for an aligned interaction between humans and AI is apparent in human-AI teaming, where an effective interaction requires
% advances in human-AI teaming.
%, especially in complex application scenarios, such as automotive driver assistance or decision-making in medicine. 
% However, effective human-AI teaming requires \textit{alignment} of their interaction properties. Alignment implies at least 
% that 

\begin{figure}
    \centering
    \includegraphics[width=1\linewidth]{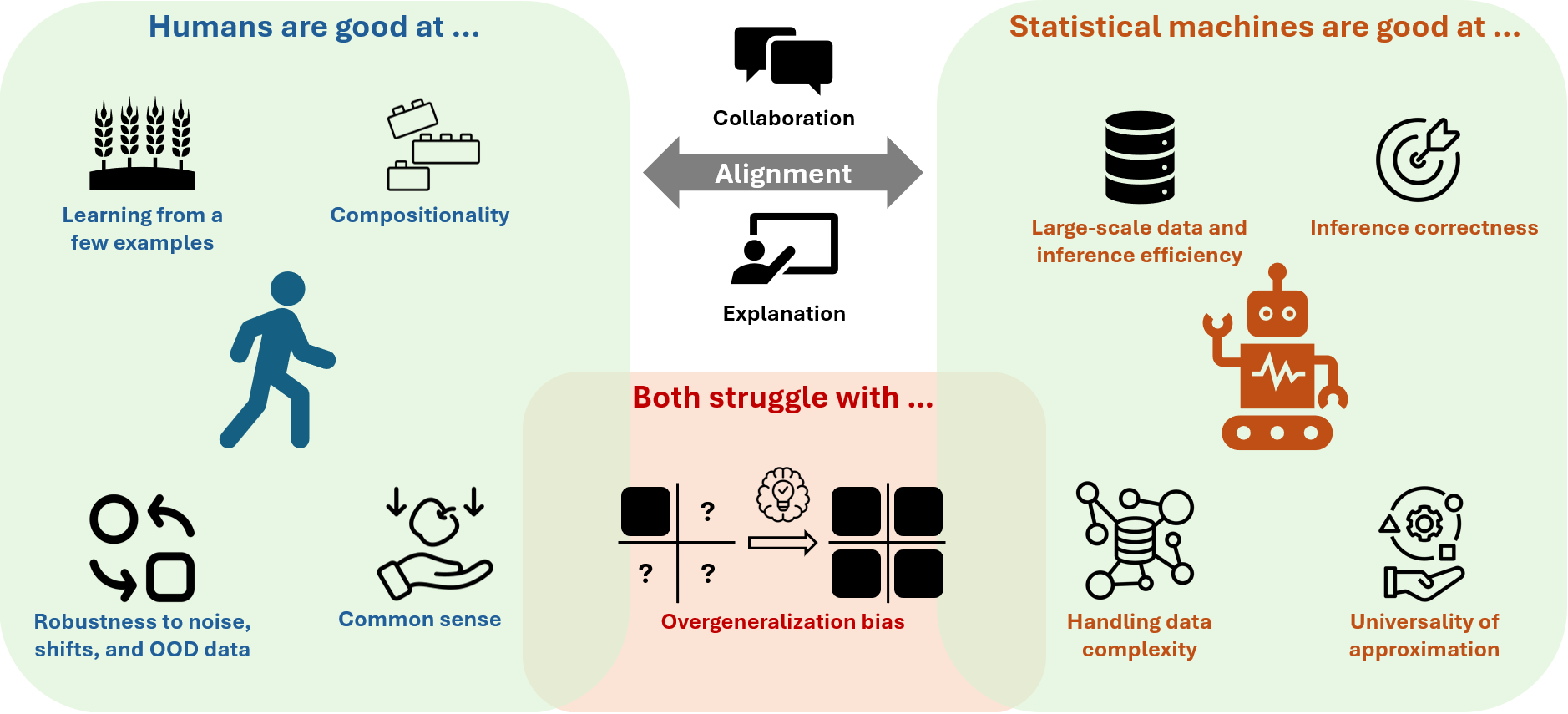}
    \caption{Comparison of the strengths of humans and statistical ML machines, illustrating the complementary ways they \textit{generalise} in human-AI teaming scenarios. 
    Humans excel at compositionality, common sense, abstraction from a few examples, and robustness. Statistical ML excels at large-scale data and inference efficiency, inference correctness, handling data complexity, and the universality of approximation. Overgeneralisation biases remain challenging for both humans and machines. 
    Collaborative and explainable mechanisms are key to achieving alignment in human-AI teaming.
    See Table~\ref{tab:summary} for a complete overview of the properties of machine methods, including instance-based and analytical machines.}
    \label{fig:human_machine}
\end{figure}

A crucial, yet often overlooked aspect of this alignment in interactive scenarios is the complementary ways in which humans and machines \textit{generalise} (Figure \ref{fig:human_machine}). 
Generalisation is typically defined as \textit{``the process of transferring knowledge or skills from specific instances or exemplars to new contexts''}~\citep{son2008simplicity}.
% \textit{a transfer of what has been learned in one context to a new, potentially similar one} \citep{goldstein2015cognitive}.
% Generalisation is typically defined as a \textit{``transfer of past learning to present situations if the conditions in the situations are regarded as similar''} \citep{gluck2013learning}.
% \textit{Generalisation is typically defined as a transfer of what has been learnt in one context to a new, potentially similar one} \citep{goldstein2015cognitive}. 
In cognitive science, human generalisation commonly involves concept learning and the abstraction of general characteristics to a collection of entities
\citep{medin2005cognitive,harnad2017cognize}.
Humans excel at generalising from a few examples, compositionality, and robust generalisation to noise, shifts, and Out-Of-Distribution (OOD) data
% reasoning about causal implications, and filling gaps in experience using abstraction, common sense, and structured knowledge 
\citep{holzinger2023toward}. 
%, even within limited information and time constraints
A major reason why humans can learn from little data and seemingly generalise beyond the observed distribution is that, through evolution, experience, or both, they have access to strong \emph{common sense} priors at multiple hierarchical levels that characterise physical principles in nature and human behaviour in interactions, often driven by causal inferences. 

 This contrasts sharply with data-driven (\textit{statistical}) AI systems, which struggle to generalise beyond their training distribution and abstract effectively. Although some neural architectures might display better alignment with physical laws \citep{Lin2017}, the generalisability of statistical machines, if averaged over all distributions and in the absence of prior knowledge, is constant (no-free-lunch theorem~\cite{wolpert1997no}).
 % \citealp{6795940}).
Unlike humans, statistical learning systems are typically driven by correlations rather than causal inference. Consequently, they excel in handling large-scale data and inference efficiency, inference correctness, handling high data complexity, and the universality of approximation.

% and, consequently, excel in data faithfulness, handling large-scale data, and high complexity.\footnote{Here, we refer to the term `data faithfulness´\ as the system's ability to accurately reflect possibly complex information as represented by the data and find regularities in high-dimensional, very large, and complex data.}

The goal of \textit{human-machine teaming} \citep{vats2024survey} is that each side addresses the limitations of the other while aligning the goals. For example, some generalisation capabilities of large language models (LLMs), like the quick production of rhetorically polished texts on any topic, are beyond those of most humans. However, their overgeneralisation errors (``hallucinations''~\cite{10.1145/3571730}), like replacing specific facts with nonfactual information, can be easily caught by a human expert.\footnote{Human overgeneralisation errors are also common, albeit of a different nature, like stereotyping~\citep{dijker1996stereotyping}.} 
Effective teaming requires that 
% humans have a mental model of AI, which implies that 
humans must be able to assess AI responses and access rationales (or ``explanations'') that underpin these responses (Figure \ref{fig:human_machine}).

The complementarity of humans and AI, and the requirements for effective human-AI teaming, shed new light on traditional \textit{analytical} (or knowledge-informed) and \textit{instance-based} AI paradigms. Analytical methods excel at compositionality and accessible semantics, albeit in limited scenarios~\citep{koller2009probabilistic}, whereas instance-based models are robust to distributional shifts provided an adequate representation is available~\citep{7837853}. Combining the strengths of various machine methods has inspired emerging research directions under the umbrella of neurosymbolic AI \citep{DBLP:series/faia/369, marcus2003algebraic}.
%Neuro-symbolic AI aims to preserve the strengths of currently dominant (neural) statistical models, such as scaling and capturing complexity, while enhancing their ability for abstraction and justification by leveraging symbolic approaches, ultimately enabling more effective human-AI teaming.

This perspective paper draws on insights about the generalisation of humans and machines from AI and cognitive science. We analyse three dimensions from the perspective of AI alignment: \textit{notions} of, \textit{methods} for, and \textit{evaluation} of generalisation. Along these dimensions, we map the different conceptualisations of generalisation in AI and cognitive science while focusing on the following questions: What are the known notions of generalisation?
%in humans and AI? 
What are the strengths and weaknesses of the generalisation of AI methods? What is the state-of-the-art of evaluating generalisation?
%the generalisation properties of AI? And finally, 
What is the impact of current trends in AI, such as foundation models, on generalisation theories, methodological frameworks, and evaluation practices?
% What are the generalisation capabilities and limitations of human-AI teams?
%regarding generalisation? 
Addressing these questions points to the need for interdisciplinary approaches to provide a basis for effective and cognitively supported alignment of human and AI generalisation.
% reveals interdisciplinary challenges that must be tackled to

\section{Parallels in Generalisation by Humans and Machines}
\label{sec_two}

% Tries to show from historical perspective the intersection between CogSci and ML research - towards approaches for generalisation

\begin{figure}[!h]
\centering

\subfloat[Learning the relational rule `grandparent' using a background theory `parent' in \citet{muggleton1994inductive}, conjunctive rules in \citet{bruner1956study}, and names of alien objects modelled as Bayesian inference over a tree-structured domain representation in \citet{tenenbaum2011grow}.]{%
  \includegraphics[width=0.8\linewidth]{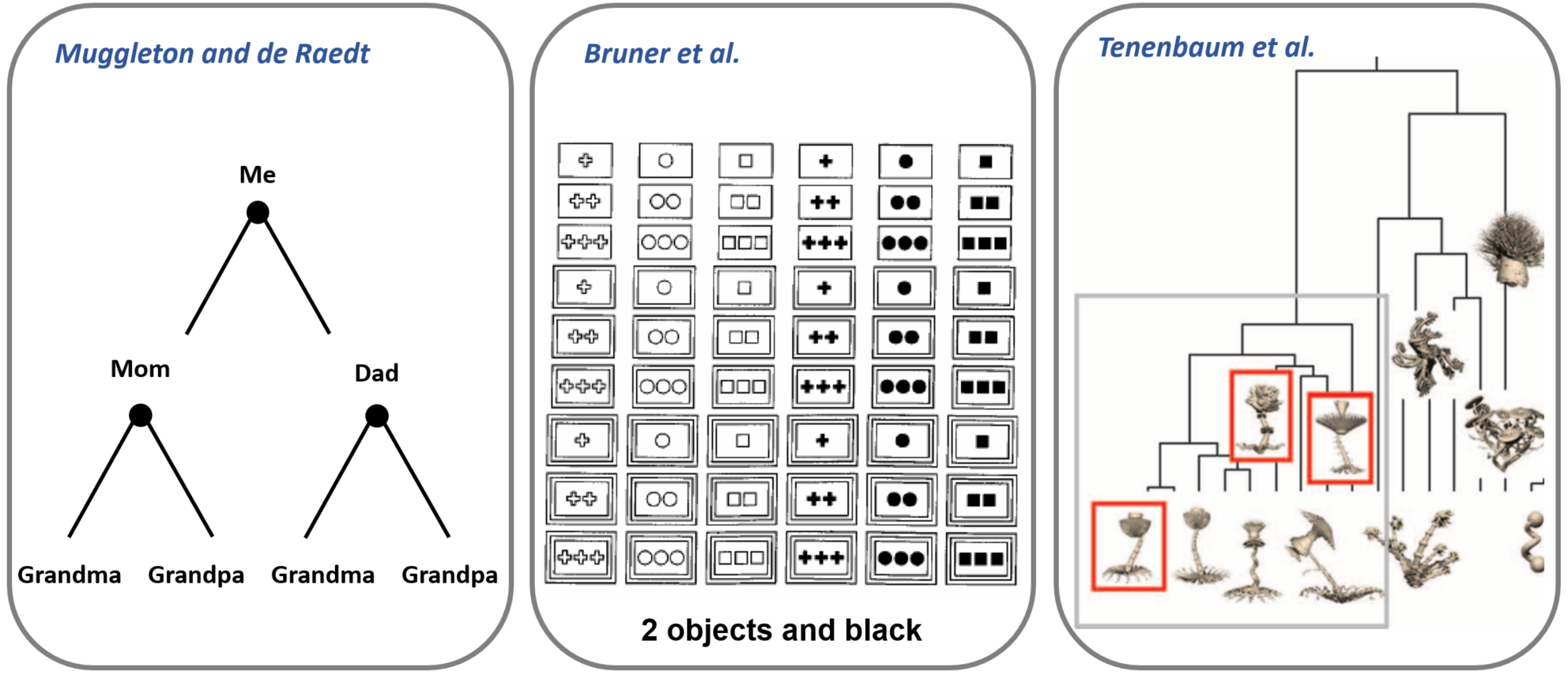}%
  \label{fig:history_rule_learning}%
}\qquad
\subfloat[Example-based prototypical representations (cf.\ \citet{rosch1975family}, and \citet{MEDIN1987242}), context-effects (\citet{labov1973boundaries}), and analogy (\citet{falkenhainer1989structure}).]{%
  \includegraphics[width=0.8\linewidth]{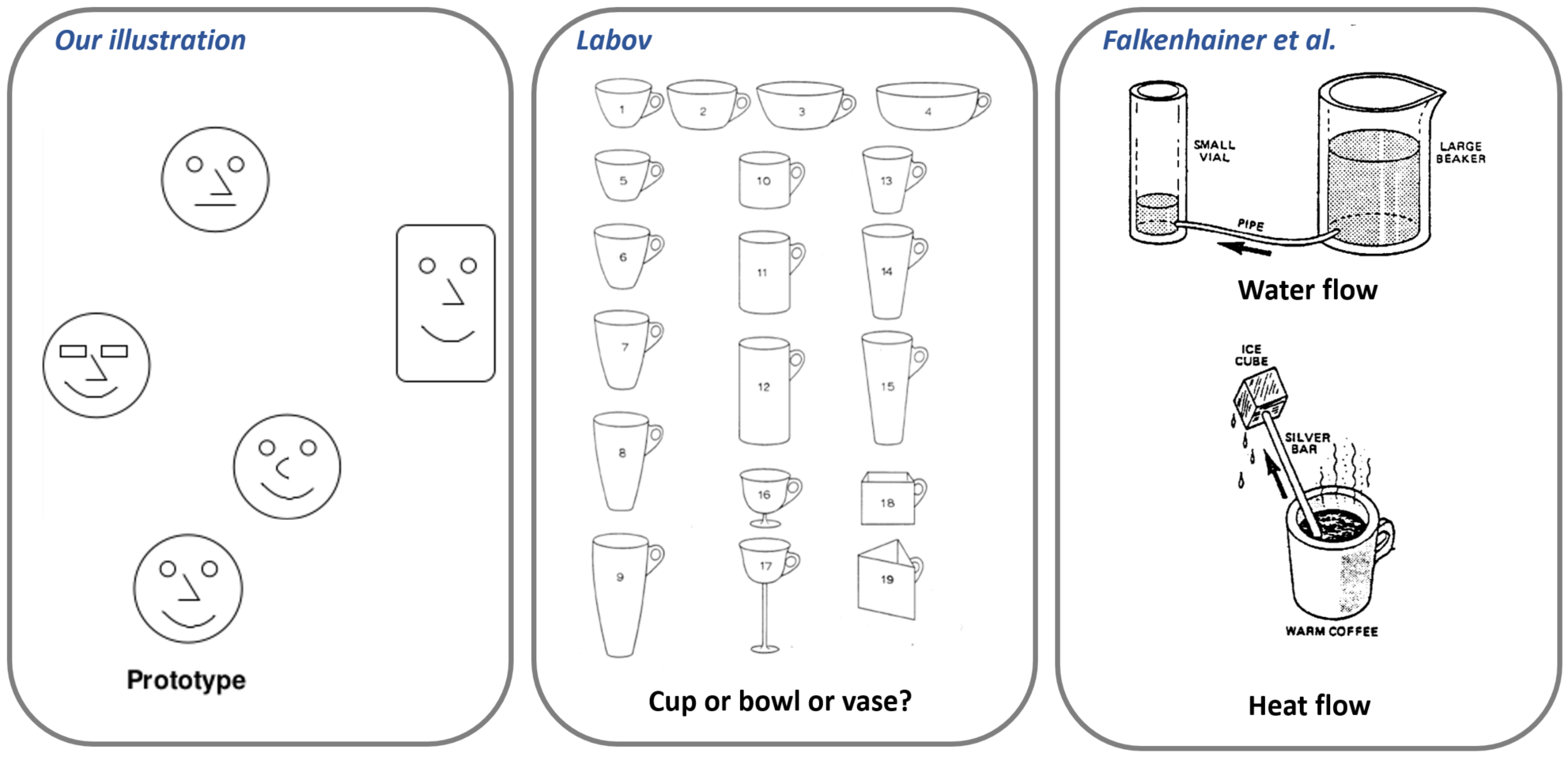}%
  \label{fig:history_prototype_learning}%
}
\qquad
\subfloat[Statistical generalisation: neural network model of semantic memory (\citet{rumelhart1986parallel}; crop of the illustration).]{%
  \includegraphics[width=0.58\linewidth]{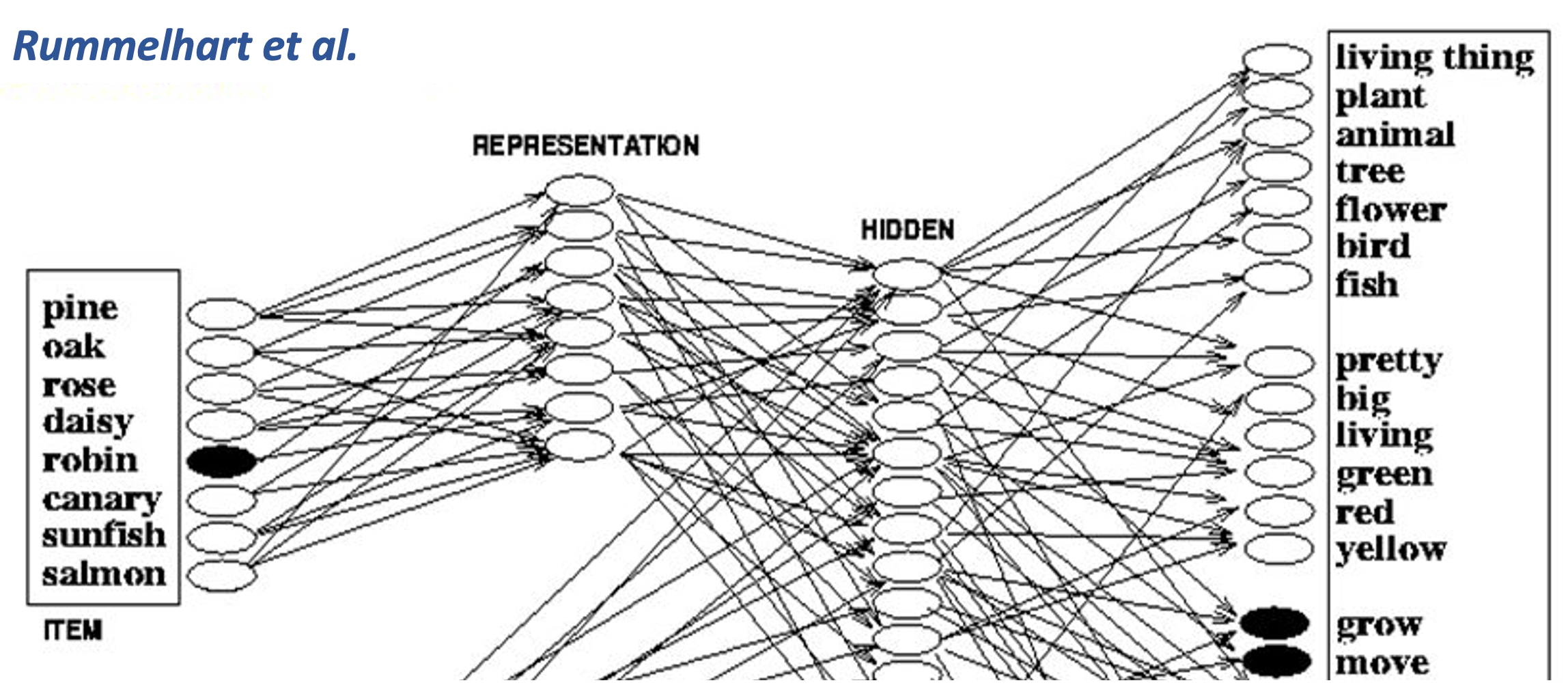}%
  \label{fig:history_statistical_learning}%
}
\caption{Illustrative examples of human generalisation and its inspiration of rule-based (top), example-based (middle), and statistical ML approaches (bottom).}
\label{fig:history}
\end{figure}

\noindent Approaches to generalisation %as an abstraction of general characteristics (features) of an entity or a concept from experience 
have been proposed in the context of AI %and Machine Learning (ML) 
as well as in cognitive psychology, and they often mutually inspired each other. This holds for all types of approaches, whether rule-based, symbolic and knowledge-informed, case- and analogy-based, as well as neural and statistical.\footnote{The generalisability of these three families of approaches will be discussed in detail in Section \ref{sec_methods}.} In the following, the mutual influence between AI methods and cognitive psychology will be illustrated by selected historical milestones, summarised in Figure~\ref{fig:history}. 

In the early days of cognitive psychology, \citet{bruner1956study} presented empirical studies investigating human concept learning (Figure~\ref{fig:history_rule_learning}),
which inspired the first decision tree learning algorithms \citep{hunt1966experiments,quinlan1987simplifying}. Observations on human learning of relational concepts \textemdash such as family relations or compound objects \textemdash inspired early machine learning (ML) approaches to learning from structural representations \citep{winston1970learning,muggleton1994inductive} and recursive concepts \citep{schmid2011inductive}. This class of approaches, often referred to as inductive programming, allows learning from a few examples and taking into account background knowledge for model induction \citep{gulwani2015inductive}. Rule learning approaches have been extended to statistical relational learning \citep{de2008probabilistic} to overcome the brittleness of purely symbolic learning. Bayesian approaches to rule learning have been introduced as a plausible framework to model human learning in complex domains such as language acquisition \citep{tenenbaum2011grow,lake2015human}.

% (2) Example-based, prototypes, analogy

%Although there is substantial empirical evidence for human concept learning as learning of rules \citep{bruner1956study,lafond2009decision}, this learning approach is typically evident only in specific settings. 
Rule learning as an explicit approach (system 2~\cite{kahneman2011thinking}) is apparent for domains of high-level cognition where relevant features and possibly relations can be verbalised \citep{lafond2009decision}. Other cognitive theories have been proposed for domains where knowledge is not (entirely) available in explicit form. 
%, and other machine learning approaches than rule learning are more suitable and effective. 
For example, prototype theory was proposed as a similarity-based approach where entities are grouped into concepts or categories for which similarity within category borders is maximised
and between categories minimised \citep{rosch1975family}, see Figure~\ref{fig:history_prototype_learning}. This approach is reflected in similarity-based methods to ML, especially k-nearest neighbours \citep{cover1967nearest}. 
%While cognitive theories address the grouping of objects into basic, super-, and subcategories---for example, empirically demonstrated by \citet{rosch1978principles}, chair as a basic category, furniture as a super-category, and office chair as a sub-category---similarity-based machine learning typically does not address the hierarchical structure of categories. 
%Another similarity-based approach proposed in cognitive research is 
Exemplar theories \citep{nosofsky1988exemplar} have been proposed to address the flexibility of human categorisation. For instance, context-dependence of the classification of visual objects has been shown by \citet{labov1973boundaries} where a cup might be classified alternatively as a bowl or a vase when different contexts, e.g., soup or flowers, are induced.
Another similarity-based approach is analogical reasoning, addressing knowledge transfer from one situation to another, often from a different domain \citep{gentner1983structure,falkenhainer1989structure}. In contrast to other methods, the analogy is not based on feature but on structural similarity \citep{gentner1997structure,wiese2008mapping}. Computational models of analogical transfer and learning \citep{gentner2011computational} address generalisation in highly complex cognitive domains such as physics. 
%In cognitive artificial intelligence, analogy-based approaches have been, for instance, proposed for problem-solving in physics \citep{forbus1983learning,gust2006metaphors} and for challenging abstract reasoning problems \citep{lovett2007analogy,clark2018think}. 
%FvH It is an open question whether large language models perform human-like zero-shot learning for solving verbal and visual analogies \citep{hu2023context,webb2023emergent,lewis2024using,sourati2024arn}.

% (3) statistical, neural

Neural network approaches, especially multilayer perceptrons, were proposed by cognitive scientists \citep{rumelhart1986parallel} as a method of generalisation learning that overcomes the brittleness of symbolic approaches; see Figure~\ref{fig:history_statistical_learning}. %Indeed, the rebirth of neural network research in machine learning at the end of the 1980s was triggered by the seminal work of McClelland and Rumelhart after a long neglect of this type of machine learning approaches mainly due to the results of Minsky and Papert \citep{minsky1969introduction}.
Despite strong arguments from researchers in symbolic AI \citep{fodor1988connectionism}, neural networks and other statistical approaches became the most dominant branch of ML due to their superior performance in increasingly large datasets. However, some core arguments of the critiques by \citet{fodor1988connectionism} concerning the relationship between statistical ML and human cognition remain. First, data are separated from a semantic model. While humans who have learnt a concept robustly recognise OOD inputs, ML has only addressed this problem more recently \citep{yang2024generalized}. Second, even if knowledge is primarily implicit in many domains, humans can verbally describe at least part of what constitutes a concept. This observation has recently been reflected in research on explainable AI, where novel approaches to explain black-box models have been proposed focusing on explanations based on concepts and relations \citep{achtibat2023attribution,finzel2024relation,DBLP:conf/nesy/DalalRBVSH24}. Third, human explanations are typically based on the causal history of an event and a causal explanation for the generalisation itself \citep{miller2019explanation}. Substantial empirical evidence has been given, demonstrating that humans do not focus on the superficial level of event covariations, but reason and learn based on deeper causal representations \citep{waldmann2006beyond}. In ML, discovering high-level causal variables from low-level observations remains a significant challenge \citep{scholkopf2021toward}.

The combination of implicit, neural learning, and explicit, symbolic approaches is addressed in neurosymbolic AI research \citep{DBLP:series/faia/369}.
% \citep{marra2024statistical,DBLP:series/faia/369,DBLP:conf/ijcai/DemirN23}. 
%Here, different architectures have been proposed, such as transforming symbolic structures into neural network models for more efficient learning (graph neural networks, \citealp{lamb2020graph}; deep Q learning over embeddings, \citealp{DBLP:conf/ijcai/DemirN23}) or using concepts learned with deep neural networks as input for rule learning systems \citep{MANHAEVE2021103504}.
Neurosymbolic approaches promise to combine the strengths of neural and analytical techniques, preserving superior performance while enabling data-model separation and integrating available background knowledge \citep{marcus2003algebraic}. Combining neural and symbolic approaches to generalisation and explainability within the so-called human-centric AI reflects many aspects of human learning flexibility~\citep{ilievski2025}.
% . It may be a promising research direction for effective human-AI teaming
For effective joint decision making and problem solving, the human-AI interface must align with human information processing \citep{ji2023ai}. In general, alignment must be established for different aspects, such as knowledge state, current information needs, or even values \citep{butlin2021ai}. In cognitive modelling, researchers aim to align the process level between algorithmic approaches to learning and human learning \citep{langley2013central}. In general, for effective human-AI teaming, it is sufficient that the output of a learnt model is aligned with human cognitive concepts. That is, the result, and not necessarily the process, of generalisation is aligned. Recent results show that, to date, the performance of human-AI teams lags behind that of the best AI model or the best human alone in many domains \citep{vaccaro2024combinations}.

%Combining strong generalisation and explainability could make neuro-symbolic systems adequate for effective human-AI teaming in many tasks.
%, including skill acquisition and problem-solving. 
%However, today's neuro-symbolic methods sometimes report lower accuracy and scalability, limiting their practical significance and requiring further research. 

%Ute: Include: representation format vs learning method

\section{Notions of Generalisation}
\label{sec:types}

In the broader context of cognitive science and AI research, there are three different notions of generalisation, which we will cover in turn.

\subsection{Generalisation as a process}
\label{sec:types_process}

Generalisation refers to a \textit{process} by which general concepts and rules are constructed from example data. In cognitive science, the process is typically called \emph{abstraction}, either through associative learning, generalisation by similarity, or the transformation of schemas from lower to higher stages of cognitive development \citep{colung2003emergence,campell2014studies}. More broadly, \citet{french1995subtlety} distinguishes three types of generalisation processes: (1) generalisation of concrete instances into an abstract schema, which we call \textit{abstraction}, (2) generalisation through the application or extension of the schema to various situations, which we call \textit{extension}, and (3) generalisation involving the transformation/adaptation of the schema to fit a new context, which we call \textit{analogy}. These three subtypes of generalisation processes are also recognisable in analytical, statistical, and instance-based AI\@. Abstraction (type 1) relates to concept learning or rule mining in analytical AI, as well as clustering/classification (for discrete classes) and regression/dimension reduction (for functional relationships) in statistical AI \citep{biehl2023shallow}. Model extension (type 2) relates to online, multi-task, few-shot, or continual learning schemes in statistical and instance-based AI methods \citep{lu2018learning,zhang2021survey,verwimp2023continual}. Analogy (type~3) relates to transfer learning \citep{zhuang2020comprehensive} in statistical AI and reasoning by analogy in analytical AI\@.

Importantly, a generalisation process does not have to start from example data but may abstract, extend, or transfer a pre-existing model beyond its original scope.

\subsection{Generalisation as a product}
\label{sec:types_product}

Generalisation refers to the \textit{products} of a generalisation process, such as categories, concepts, rules, and models, in their various representations \citep{reilly2003we}.
Generalisations of \textit{categories and concepts} may be represented as a symbolic definition, as a list of attributes and their bounds (refer to \citep{jackendoff1985semantics} for the cognitive science perspective, and \citep{bertsimas2017optimal} for decision trees in AI), as a prototype (refer to \citep{mervis1981categorization} for a cognitive science, and to \citep{bien2011prototype} for an AI perspective), or as a set of exemplars of the category (refer to \citep{nosofsky2011generalized} for a cognitive science perspective, and to \citep{peterson2009k} for the k-nearest neighbours scheme in AI). In probabilistic models, categories or concepts are represented as a probability distribution from which examples of this category can be drawn, which is also the notion implicitly used by generative AI models \citep{bengesi2024advancements}.
Beyond categories or concepts, products may also be \textit{rules or relations}, represented, for instance, via functions or graphs in parametric or non-parametric form.

%Whether we consider concepts, categories, rules, or relations, a core purpose of generalisations is the successful application to new data.

\subsection{Generalisation as an operator}
\label{sec:types_data}
The purpose of generalisation (as a \textit{process}) that produces a generalisation (as a \textit{product}) is to apply the generalisation (as an \textit{operator}) to new data. 
The ability of a model to generate accurate predictions on new data is at the core of generalisation in statistical AI \citep{adams2022controlling,shalev2014understanding}.
A classical assumption of statistical AI is that data are Independent and Identically Distributed (IID), that is, data for training resemble data for generalisation.
Three mathematical theories of generalisation have emerged in the literature: (1) The \textit{Probably Approximately Correct (PAC)} framework analyses whether a model (\ie a product) derived via a machine learning algorithm (\ie a generalisation process) from a random sample of data can be expected to achieve a low prediction error on new data from the same distribution in most cases \citep{shalev2014understanding}. (2) \textit{Statistical physics of learning} aims to understand the typical properties of learning algorithms (\ie processes) with many adaptive parameters \citep{decelle2022introduction, EB01, PhysRevA.45.6056}. (3) \textit{Vapnik--Chervonenkis (VC) dimension theory} focuses on the storage capacity of model classes and their subsequent ability to make accurate predictions on new data \citep{vapnik2015uniform}. One of the key insights in this context is that generalisation begins where memorisation ends, paraphrasing \citet{cover1965geometrical}. In other words, only if the capacity of a system to memorise is limited or restricted, can it generalise to novel data. 
In contrast, humans can generalise from a few samples for a specific task, as generalisation in humans is not a singular event but based on lifelong experience of regularities observed in nature. Few-shot learning in machines mimics some aspects, such as the transferability
of learnt representations across tasks
\citep{muller-etal-2022-shot}.

These formal theories cover the relevant setting of model generalisability where training and test data are IID\@. The case of OOD generalisation, which means that the test data are different from the data used for training, is covered only to a minimal extent, and no guarantees are available in the general case
\citep{straat2022supervised}.
Importantly, all three theories have been applied mainly to processes of the abstraction kind (subtype 1 above). For model extension or analogy, the mathematical theory is less well-established \citep{liu2021towards,straat2022supervised,tripuraneni2020theory}. Of particular interest is generalisation across compositions \textemdash for instance, in language \textemdash which has been addressed in analytical AI \citep{GOLD1967447}, but is limited by the undecidability or high complexity of their inference \citep{Zeugmann}.
Despite such fundamental restrictions, humans operate with compositions similar to those found in language and vision. Key aspects which ensure feasibility are regularisations due to multi-modality and natural priors, which restrict the otherwise combinatorial exhaustive space of possible models \citep{Dingemanse2015}.

%Beyond machine learning theory, the notion of similarity appears crucial: the more similar new data is to the data the original model was constructed from, the more straightforward generalisation becomes. In abstraction, new data stems from the same source; in extension, new data tends to be closely related; but in analogy, data may stem from entirely different domains, and similarity to the original model first needs to be established (by analogy, a structured mapping, or some other mechanism). 

\subsection{Alignment of human and machine notions of generalisation} 

We observe that human and machine notions of generalisation are misaligned. While humans tend toward sparse abstractions and conceptual representations that can be composed or transferred to new domains via analogical reasoning, generalisations in statistical AI tend to be statistical patterns and probability distributions, which can sometimes be extended but still fail to generalise to tasks and domains that are too far removed from the training data. In other words, because humans and machines use different \textit{processes} (\eg abstraction vs.\ data-driven learning), they arrive at different \textit{products} (\eg categories and rules vs.\ probability distributions) that generalise differently; if we wish to align machines to human-like generalisation ability (as an \textit{operator}), we need new methods to achieve machine generalisation.
\section{Machine Methods for Generalisation}
\label{sec_methods}
\begin{table}
\caption{Common categories to structure AI methods algorithmically centred.
These categories are not uniquely related to their type of generalisation.\label{tab:AI_methods}}
\begin{tabular*}{1\textwidth}{@{\extracolsep{\fill}}>{\centering}p{0.21\textwidth}>{\raggedright}p{0.79\textwidth}}
\toprule 
\textbf{Category} & \textbf{Attributes}\tabularnewline
\midrule
\midrule 
Training signal & supervised, unsupervised, reinforcement,
semi-supervised,
self-supervised\tabularnewline
\midrule 
Data type & tabular data, data structures (\eg text, graph),
prior knowledge\tabularnewline
\midrule 
Model representation & parametric/non-parametric, symbolic/sub-symbolic,
black-/white-/grey-box\tabularnewline
\midrule 
Training objective & Bayesian inference, maximum likelihood principle,
rule learning, mean \\
& squared error minimization\tabularnewline
\bottomrule
\end{tabular*}
\end{table} 
As elaborated in Section~\ref{sec_two}, 
humans excel in
systematic generalisation across representations, contexts, and tasks based only on a few observations. Which computational methods can achieve such generalisation capabilities?
Although recent work has shown remarkable results for specific neural networks in solving compositional tasks \citep{Lake2023}, the underlying mechanisms of human generalisation are not understood sufficiently to mimic them in artificial systems \citep{Tenenbaum_Griffiths_2001}.

AI methods are usually structured according to algorithmic aspects rather than their generalisability
(see Table~\ref{tab:AI_methods}).
Although algorithmic choices impact generalisation behaviour \textemdash for instance, symbolic methods often implement compositionality \textemdash there is no simple mapping from these algorithmic aspects to the form of generalisation that an AI model can achieve. 
Therefore, we focus on another categorisation, the interplay of observational data (\ie single instances) and models (\ie principles that apply to a whole population) in the following, as this correlates to the generalisability of the model and its suitable evaluation (Section~\ref{sec:evaluation}). Three categories can be distinguished: (1) The transfer of individual observations to a population is the basis for \emph{statistical generalisation methods}. (2) The search for observational evidence of an explicit model or theory is done in  \emph{knowledge-informed methods}. (3)~\emph{Instance-based methods} such as case-based reasoning or nearest-neighbour methods, focus on individuals concerning the source and target of generalisation.
These choices have different characteristics in terms of their generalisability and alignment with human generalisation.

\subsection{Statistical generalisation methods in AI}
%ML method as empirical loss optimization, regularization ensures generalisation
Many modern ML methods, including deep learning, aim at statistical generalisation: observational data (\ie training data) serve as input to an inference mechanism that extracts a model for the entire population (\ie the underlying distribution).
Generalisation refers to the ability of the inferred patterns to be successfully applied to new data (Section~\ref{sec:types_data}).
Typically, algorithmic methods are expressed as optimisation methods for a model's loss function, such as the model prediction error.
As the loss cannot be evaluated in the entire population, it is approximated in a given training set, known as \emph{empirical risk minimisation} \citep{vapnik95}.
Although evaluation in an independent test set constitutes an unbiased estimator of generalisation ability (Section~\ref{sec:evaluation}), the empirical loss in the training set systematically underestimates the loss of the model.
Therefore, \emph{generalisation needs to be accounted for explicitly}.
 Popular strategies include regularisation terms that favour models with better generalisation behaviour, such as maximisation of margin or stability \citep{journals/jmlr/BousquetE02,DBLP:journals/neco/SchneiderBH09a}. 
Even heavily over-parametrised deep learning models can lead to surprising generalisation capabilities due to intrinsic regularisation \citep{Grohs_Kutyniok_2022}.

Statistical methods generally excel in inference correctness and efficiency for large-data sets.
Yet, powerful models also typically require large data sets for successful training.
This challenge can be partially overcome by technologies that do not learn from scratch but build their inference on already learnt representations and principles of instance-based translations,
such as few- or zero-shot mechanisms \citep{ge2023few,muller-etal-2022-shot}.
Moreover, empirical risk minimisation has a fundamental limitation compared to human generalisation:
\emph{generalisation can only be expected in areas
covered by observations}, but not for out-of-sample events, novel contexts, or distributional changes \citep{ye2021towards}.
Indeed, machine behaviour for OOD settings can significantly deviate from human expectation, with adversarial attacks as prominent examples of this phenomenon
\citep{7467366}.

In recent years, many vital settings, including generative language models, have been targeted that do not allow for a simple analytic loss function that fully describes human intentions.
Thus, surrogate losses, such as the cross-entropy for next token prediction, are used as a proxy.
With massive training data, instruction tuning, or human feedback, impressive generalisability arises \citep{10.5555/3495724.3495883}.
However, \emph{the emerging generalisation abilities are only partially understood and do not necessarily align with human expectation}, necessitating a downstream evaluation (Section~\ref{sec:evaluation})
\citep{bommasani2022opportunitiesrisksfoundationmodels} if possible at all. Indeed, human intentions are not necessarily well-formed or static, and the type of information an AI provides could influence human objectives in interactive scenarios
\citep{10.5555/3692070.3692292}. Therefore, modelling the uncertainties and partial misalignment of these surrogate losses in AI systems is necessary.

Statistical generalisation methods are often based on model families with universal approximation capability to account for the lack of domain-specific knowledge. 
As a consequence, deep models, for example, can deal with high degrees of nonlinearity and complex functions, including multimodal
signals \citep{JIAO20241}. 
Yet, the product is typically a black box, which does not reveal insight into its generalisation behaviour; moreover, as universal families are very powerful, partially unintended generalisation behaviour can easily occur. 
Recently developed post hoc \emph{explanation methods} allow for a closer inspection of the underlying rationale and its impact on the generalisation behaviour of the model
\citep{DBLP:conf/nesy/DalalRBVSH24}. 

\subsection{Knowledge-informed generalisation methods in AI}
Knowledge-informed generalisation methods aim to find empirical evidence of a theory. The resulting product is a meaningful representation confirmed by the data. Various methods exist, such as mechanistic models  \citep{Baker2018-la}, causal models  \citep{10.1145/3444944},  or functional programs \citep{10.5555/1248547.1248562}. 
As the semantics of the model is directly accessible, \emph{humans can inspect how these models generalise to previously not encountered scenarios}, which means that model predictions are explainable by design, and the generalisation of the model is often well aligned with human expectations. 
Yet, model parameters require semantic grounding, challenging to realise with subsymbolic, low-level sensor data. 
Neurosymbolic integration can partially overcome such limitations \citep{MANHAEVE2021103504}. 

\emph{Learning the optimal model structure is demanding}, both numerically and due to fundamental limitations such as non-identifiability of structural components \citep{koller2009probabilistic}. Hence, many methods are restricted to simple schemes, such as description logic, rather than universal approximators. Learning methods are diverse because they mirror specific representations; examples include semantic clustering, probabilistic rule mining, subsumption, and analogies. Since noise robustness and inference efficiency pose significant challenges, hybrid approaches have emerged, such as embedding mechanisms, which transfer a symbolic model to a real-valued embedding space where efficient numeric inference is possible \citep{10.1145/3643806}.
Knowledge-based approaches enable the explicit inspection and manipulation of knowledge, and they allow generalisation based on a few examples, as these can be enhanced by explicit rules that ensure valid generalisation. Yet, these approaches are restricted to domains where a theory can be formalised with reasonable effort. As this is often limited, currently, such models do not reach the impressive capabilities of statistical approaches trained on massive datasets such as LLMs.

\emph{Systematic compositionality} refers to the ability to generalise and produce novel combinations from known components. It has been fundamental in the design of traditional, logic-based systems; yet statistical methods have struggled with compositional generalisation \citep{fodor1988connectionism}. 
Compositionality seems to be a universal principle in nature, since it has been observed not only in humans but also in many other species \citep{doi:10.1126/science.adv1170}.
In recent years, significant progress has been made in improving compositional generalisation in deep learning, typically by adding analytical components that mirror the compositional structure of the domain, such as structure-processing neural networks \citep{10.5555/518982} or metalearning for compositional generalisation \citep{Lake2023}. Although these efforts provide a pathway for neural networks to generalise systematically, most of the results are empirical, making achieving predictable and systematic generalisation challenging \citep{wiedemer2024provable}. There remains a \emph{significant gap between the systematic generalisation capabilities of knowledge-informed models and the representation learning techniques of deep models}, with neurosymbolic AI promising a viable bridge \citep{DBLP:conf/iclr/SehgalGSC24}.

\subsection{Instance-based translation in AI}
Lazy learning methods refer to non-parametric techniques, such as nearest neighbours methods or case-based reasoning \citep{aha2013lazy}. They rely on local inference, computed when needed based on similar cases encountered previously. The Nearest-neighbour methods are among the most popular ML methods, showing high flexibility when combined with complex representations \citep{Khandelwal2020Generalization}. 
Since non-parametric methods adjust their complexity as needed, they offer universal approximation capability. Yet training and inference are also efficient in large data sets, given the availability of suitable data structures. As instance-based methods rely on local inference, human inspection of individual decisions \textemdash although not of the entire model \textemdash is usually possible.
Instance-based and exemplar-based methods closely resemble concepts in cognitive linguistics, such as a graded degree of belonging to a category, which can be represented by a prototype \citep{ROSCH1973328}.

Due to their local inference and explicit memory, instance-based models have shown great promise for incremental learning of distributional shifts \citep{7837853}. 
Since they memorise single instances, they can identify
out-of-sample instances based on their similarity to previously encountered data. Moreover, they can naturally deal with the challenge of catastrophic forgetting in continual or \emph{lifelong learning} as they can keep possibly relevant data points due to the
explicit and local representation of information. 
This principle also suggests possible solutions to
catastrophic forgetting in continual learning using deep statistical models
\citep{NEURIPS2020_ca4b5656}.
Conversely, it is possible to implement an explicit
forget mechanism in the case of real concept drift, which means older instances become invalid.
The reliance of instance-based methods on similarity means that \emph{a suitable representation is key to support generalisation}  \citep{MAL-019}, as it directly influences the model's ability to evolve patterns across diverse datasets and tasks. Recent work investigates how to achieve representations to support generalisations across tasks or domains \citep{he2022learning}.

\emph{Context has a unique role} as {generalisation requires adapting knowledge learnt in one setting to fit a novel, unseen one}. Humans can cope with the challenge of acquiring and representing context knowledge, as well as assessing the similarity of two contextual representations \citep{doi:10.1126/science.3629243}. ML techniques such as transfer learning, prompting, or retrieval augmented generalisation mimic parts of this process \citep{gao2024retrievalaugmentedgenerationlargelanguage}.  In this realm, LLMs have demonstrated remarkable capabilities for few-shot or in-context learning \citep{10.5555/3495724.3495883}, still often with only implicit contextual information. An explicit representation of contextual knowledge, for example, through neurosymbolic AI, is the subject of ongoing research.

\subsection{Aligning machine generalisation methods and human expectations}

While \textit{statistical methods} constitute powerful techniques for universal approximation, their generalisation behaviour does not match human generalisation well, lacking the ability to generalise to OOD samples and exhibit compositionality. Another challenge is their black-box nature, where post hoc explanations provide solutions for specific cases. In contrast, \emph{knowledge-based methods} enable human insight and compositionality by design, but often at the expense of universality and algorithmic efficiency in the face of structure learning. Emerging neurosymbolic approaches aim to combine these methodological principles and their advantages.
\emph{Instance-based methods} try a different approach, abandoning global modelling altogether and focusing on generalisation from and to individual instances. While this underlying principle is well aligned with human generalisation and enables learning from a few data points and lifelong learning scenarios, the specific result depends strongly on the choice of representation and context.
Here, recent statistical methods for representation learning offer promising directions that allow machines to generalise from a few examples, similar to humans. Ultimately, making claims about the generalisation properties of various machine families of approaches requires a meaningful evaluation, which we discuss next.
\begin{table}
\caption{Characterization of AI generalisation methods.\label{tab:AI_methods_generalisation}}
\begin{tabular*}{1\textwidth}{@{\extracolsep{\fill}}>{\centering}p{0.5\textwidth}>{\centering}p{0.5\textwidth}}
\toprule 
\textbf{Pros} & \textbf{Cons}\tabularnewline
\midrule
\midrule 
\multicolumn{2}{c}{\emph{Statistical: generalisation from observations to a population}}\tabularnewline
\midrule 
universal approximation, surprising \\
generalisation of deep models & black boxes, generalisation only \\
within observed distribution\tabularnewline
\midrule 
\multicolumn{2}{c}{\emph{Knowledge-informed: confirm/adapt hypothesis based on observations}}\tabularnewline
\midrule 
meaningful models, identifiable \\
parameters, generalisation in the
limit, compositionality & restriction to simple scenarios, optimization/structure \\
identification computationally demanding\tabularnewline
\midrule 
\multicolumn{2}{c}{\emph{Instance-based: translation from previous observations to a new observation}}\tabularnewline
\midrule 
flexible to change/distributional shift & rely on suitable representation\tabularnewline
\bottomrule
\end{tabular*}
\end{table}

\section{Evaluation of Generalisation}
\label{sec:evaluation}

% 5185 -> 4793 (after 5.3)

% \subsection{Theory}
% \label{sec:evaluation:theory}

The theoretical generalisation strengths and weaknesses of the machine method families are summarised in Table \ref{tab:AI_methods_generalisation}. Statistical methods enable \textit{universality of approximation} and \textit{inference correctness}, and excel at handling \textit{data complexity}, \textit{large-scale data, and inference efficiency}. Analytical methods support \textit{compositionality}, \textit{explainable predictions}, and perform \textit{explicit knowledge manipulation}. Instance-based methods support \textit{robustness to noise, shifts, and OOD data}, \textit{ memorise training samples} reliably, and can \textit{learn from a few samples}.

Deriving provable robustness guarantees and generalisation bounds is necessary to define the theoretical limits of models.
Meanwhile, empirically evaluating the machine's generalisation abilities is also desirable.
% What is the theory
From a statistical learning perspective, evaluating the generalisation of supervised approaches estimates
% how well they perform on unseen data samples. 
% When testing generalisation as 
their applicability to new data (Section \ref{sec:types_data}). This formalisation of measuring \textit{inference correctness}, where training and test data samples are assumed to be independently generated by the same distribution, is theoretically grounded and remains relevant when evaluating and certifying systems. This procedure allows measuring the \textit{universality of approximation} by controlling the increase in task complexity and the ability to handle \textit{large-scale data and perform efficient inference} on larger and more complex datasets.
However, with the increasing complexity of the task and opaqueness of the system, it becomes challenging to guarantee the validity of the generalisability assumptions of the identical data distribution, as well as the representative and mutually independent sampling between the train and test sets. As an illustration, \citet{li2024task} discovered that ChatGPT performs surprisingly well on all benchmarks published before its release and much worse on all benchmarks published later. 
In this case, the independence assumption is not fulfilled as the test set information has been leaked into ChatGPT, invalidating the estimation of the generalisation performance on benchmarks published before its release. 
The lack of transparency about the data used to train LLMs makes it challenging to create novel test sets, leading to a possible overestimation of model generalisation \citep{shin2024large}. 
Namely, while the emergence of foundation models have made evaluation of \textit{learning from a few examples} via zero- and few-shot tests possible for various domains such as medicine~\citep{ge2023few}, the risk of memorisation means that test data may appear partially in their training set, invalidating the findings (data contamination~\citep{dodge-etal-2021-documenting}).
% and models that may 
% % define test data samples unseen during training. Consequently, the estimation of generalisation performance might be inaccurate so that models might 
% fail surprisingly on simple tasks 
% Further investigations are necessary to understand the theoretical limits of such models by deriving provable guarantees and properties such as generalisation bounds and provable robustness guarantees so that a model assessment beyond empirical evaluations is possible. 

The following discusses several areas related to the evaluation of generalisation and its role in recent AI applications.

\subsection{Measuring distributional shifts}

Distribution shifts can be estimated using statistical distance measures such as the Kullback-Leibler divergence or the Wasserstein distance between the feature distributions of the training and test sets \citep{liu2021towards}.
% Assessing whether empirically gathered data stem from the same distribution is non-trivial. To measure the extent to which existing datasets are OOD, statistical distance measures such as the Kullback--Leibler divergence or Wasserstein distance can quantify the divergence between the feature distributions of the training and test sets.
Generative models produce an explicit likelihood estimate $p(x)$ that indicates how typical a sample is to the training distribution. Since discriminative models do not offer this possibility, proxy techniques include calculating cosine similarity between embedding vectors and using nearest-neighbour distances in a transformed feature space. In the case of LLMs, a standard proxy for familiarity is to measure perplexity.
% Perplexity can be especially insightful if direct access to the model's internal representations and predictions is available. %This provides a quantitative measure of generalisation by evaluating how the model handles language patterns both within and outside its training corpus.
When the model's internal representations cannot be directly accessed, the layers of non-linear abstractions in modern (deep) machine learning models allow for gauging relations through intermittent embeddings.
% probing its input sensitivity is informative. For example,  
% In turn, this enables us to deepen evaluation beyond predictions or loss values and further examine changes in, e.g., representations and understand where potential discrepancies arise.
% Adversarial techniques can identify or generate data samples with distributional shifts to understand the model's robustness. 
Learning evaluation in the context of \textit{drift} can be done using tailored benchmarks
\citep{Cossu2025DontDA}.
The model's \textit{robustness to noise, distributional shifts, and OOD data} can be studied using adversarial and counterfactual techniques.
Adversarial techniques alter key data features such as syntax, semantics, or context while preserving the underlying task and the original label~\citep{jia-liang-2017-adversarial}. In contrast, counterfactual techniques create data samples that alter target prediction with minimal input changes~\citep{lewis2024using}. 

% we have a recent survey paper where a dedicated section refer to benchmark data sets for learning with drift, which might be cited here as a reference for how to evaluate learning in the context of drift https://www.esann.org/sites/default/files/proceedings/2025/ES2025-23.pdf

% To assess the generalisation capabilities of machine learning models, it is crucial to create OOD test sets and measure the extent to which existing datasets are out-of-distribution. This involves identifying or generating data samples that deviate from the distribution of the training dataset, thus representing new, unseen scenarios.
% This can be achieved, for example, by altering key features of the data—such as syntax, semantics, or context—while preserving the underlying task.
% Techniques like counterfactual data augmentation, where minimal changes are made to input data to alter its label or meaning, are effective.
% Additionally, adversarial testing, where inputs designed to confuse the model are introduced, can help.
% This approach tests the model's robustness and helps understand the limitations of current datasets in covering the problem space.

\subsection{Determining under- and overgeneralisation}

AI models are created to provide value to human users. Therefore, models are assessed relative to human generalisation abilities. In this regard, the human-centric concepts of under- and overgeneralisation are commonly used.
\emph{Undergeneralisation} occurs when a change in the input, perceptible or imperceptible, causes a considerable modification within a model. Examples of undergeneralisation include model performance degradation for a wide range of potential natural changes, such as frequent camera and environment perturbations in computer vision \citep{Hendrycks2019BenchmarkingRobustness}, since models are not invariant to these changes.
 For foundation models, prompt choice substantially affects performance \citep{gonen-etal-2023-demystifying}.
% The performance of models is known to degrade for a wide range of potential natural changes, such as frequent camera and environment perturbations in computer vision \citep{Hendrycks2019BenchmarkingRobustness}, because the model is not invariant to these changes. Such examples reveal \emph{under-generalisation} because they introduce an imperceptible shift in input that results in a considerable modification within a model. For foundation models, even the prompt choice can substantially affect performance \citep{gonen-etal-2023-demystifying}.
In contrast, models \emph{overgeneralise}, which means that they over-confidently make false predictions for (known or novel) concepts precisely because critical differences are ignored in prediction \citep{Boult2019LearningUnknown}. 
A well-known overgeneralisation phenomenon is hallucination, which refers to models that deviate from the source of the information, typically the pretraining data for LLMs \citep{10.1145/3571730}. Other inappropriate overgeneralisations are biased predictions, for example, when a model predicts a property of an individual from the statistical properties of a demographic group to which they belong \citep{hovy-spruit-2016-social}, and logical fallacies \citep{sourati2023robust}. 

% This term originally referred to models deviating from the source, such as the input document in summarization. In LLMs, the source covers pre-training data in a more general way \citep{10.1145/3571730}, extending the definition of hallucination to include factually incorrect statements. 
% Other examples of inappropriate over-generalisation are biased predictions, e.g., when a model predicts a property of an individual from the statistical properties of a demographic group to which they belong \citep{hovy-spruit-2016-social}, and logical fallacies \citep{sourati2023robust}. 

Characterising the model's under- and overgeneralisation requires choosing an appropriate metric, defining its use, and establishing a mechanism to interpret the metric's score in terms of generalisability beyond the particular test examples. The procedure is susceptible to three caveats. 
%First, the choice between discriminative and generative models, in the mathematical sense of modelling $p(y|x)$ and $p(y|x)p(x)$, determines which representational basis is used to infer similarity \citep{Mundt2023wholistic}. 
First, the choice between discriminative and generative models determines which representational basis is used to infer similarity \citep{Mundt2023wholistic}. For instance, if the task is visual classification, a discriminative model will only learn representations useful to optimize classification accuracy \textemdash in the mathematical sense it learns to directly predict labels based on data: $p(y|x)$ \textemdash whereas a generative approach would additionally learn representations necessary to describe the full data distribution \textemdash in the mathematical sense classify based on the joint probability: $p(x,y) = p(y|x)p(x)$. 
%Second, deep models are prone to learn various decision shortcuts \citep{Lapuschkin2019CleverHans} and to ignore meaningful features (simplicity bias) \citep{Shay2020SimplicityBias}. For example, a visual classification model that distinguishes oranges from avocados may learn to rely only on colour features. 
Second, deep models are prone to learning various decision shortcuts and ignoring meaningful features \citep{Lapuschkin2019CleverHans}. For example, a visual classification model could distinguish oranges from avocados by learning to rely only on colour features, even if other features could be equally descriptive of the task, sometimes also referred to as simplicity bias \citep{Shay2020SimplicityBias}. 
Third, modern models are often proprietary and frequently updated, partly based on user interactions through reinforcement learning, which increases their exposure to datasets and hinders reproducibility.
To protect against these caveats, besides using open-source models, it is critical to evaluate across different levels of abstraction, from surface forms to semantic similarity and higher-level structural mappings, and explicitly consider the application context and limits. Consequently, machines are increasingly tested for their ability to \textit{handle complex data}, such as multimodal datasets~\citep{10.1145/3713070}, and to exhibit \textit{compositionality} in tasks such as reference games~\citep{Andreas2019MeasuringCI}, commonsense reasoning \citep{davis2023benchmarks}, analogies \citep{sourati2024arn}, and concept induction \citep{nie2020bongard}.

\subsection{Distinguishing memorisation and generalisation}
% \textit{Memorisation} occurs when a model learns noise or nongeneralisable details from the training data, which do not apply to new data.
In AI, \textit{memorisation} refers to learning details from the training data, including facts and noise. %, that do not generalise to new data.
% In , 
Memorisation may be beneficial in some cases (e.\,g., Paris is the capital of France), but detrimental in others (e.\,g., Biden is the president of the United States).
This observation raises a natural question: \textit{In which cases \textit{should} models generalise, when \textit{should} they memorise, and how can this distinction be convincingly evaluated?}
The expectation of whether the models should generalise or memorise is set a priori.
When learning from experience, generalisation is crucial, for instance, in recognising a new manifestation of a vase \citep{labov1973boundaries}. 
Consequently, generalisation setups include cross-domain validation and robustness testing. 
In contrast, factual knowledge is often memorised: Paris is the capital of France, and mosquitoes fly. 
Tasks such as answering factual questions~\citep{wang2023survey} and reasoning about legal precedents~\citep{guha2023legalbench} require \textit{memorisation of training samples}.\footnote{For a review of memorisation in machine learning, see \citep{usynin2024memorisation}.} While evaluating memorisation and generalisation separately is informative, many tasks, including causal reasoning~\citep{nogueira2022methods}, argumentation~\citep{atkinson2017towards}, and theorem proving~\citep{yang2023leandojo}, require holistic integration of generalisation and memorisation, centred around \textit{explicit knowledge manipulation}.
The explicit use of knowledge enables testing the machine's ability to provide \textit{explainable predictions} using human studies and faithfulness evaluation~\citep{10.1145/3583558}.

% \subsection{Aligning evaluation of generalisation in humans and machines}
% \subsection{Significance of AI generalisation evaluation for human-AI alignment}
\subsection{Alignment of machine evaluation of generalisation to humans}

Effective alignment of AI with humans requires a principled and meaningful evaluation of its strengths and weaknesses in generalisation. Evaluating AI generalisability in the context of its alignment comprises: (1) deriving provable guarantees and limits about AI's properties, like compositionality and learning from a few samples, and (2) performing empirical evaluation by leveraging task-specific benchmarks and metrics. The evaluation of AI generalisability measures its performance concerning distribution shifts, determining its undergeneralisation (adaptability to task variations such as camera perturbations) and overgeneralisation (e.\,g., hallucinations), and its ability to memorise and generalise when necessary adequately. These three aspects are essential for alignment, since distributional shifts, task variations, and both facts and noise are natural in human-AI teaming scenarios. We summarise typical approaches for evaluating key generalisation properties in Table \ref{tab:summary}. In the next section, we discuss open challenges for measuring generalisation in the context of human-AI alignment.

\begin{table}[!h]

\caption{Desired properties of generalisation that emerge from the notions and methods of generalisation (Section \ref{sec:types} and \ref{sec_methods}). The properties are listed in order of their appearance in Section \ref{sec:evaluation}, split by a subsection using a horizontal line.
Each of the properties is supported by statistical (S), analytical (A), and instance-based
(I) AI methods, as discussed in Section \ref{sec_methods}. Achieved is indicated by "$+$" and not achieved by "$-$"; we use a strict evaluation to avoid partial scoring. The evaluation methods for each property are discussed and substantiated by relevant references in Section \ref{sec:evaluation}.  \label{tab:summary}}

\centering{}%
\begin{tabular*}{1\textwidth}{@{\extracolsep{\fill}}ccccc}
\toprule 
\multirow{2}{*}{Property} & \multicolumn{3}{c}{Method} & \multirow{2}{*}{Evaluation}\tabularnewline
 & S & A & I & \tabularnewline
\midrule
\midrule 
% 3.3, 4.1, 5: 
inference correctness & + & -- & -- & train-test splits\tabularnewline 
% 4.1, 5: 
universality of approximation & + & -- & + & increasing task complexity \tabularnewline
% 4.1, 5: 
large-scale data and inference efficiency & + & -- & + & large/complex datasets\tabularnewline
% 4, 5: 
learning from a few samples & -- & + & + & zero/few-shot tests\tabularnewline \hline
% 3.3, 4.3, 5.1: 
robustness to noise, shifts, OOD data & -- & --  & + & adversarial, shifted, counterfactual tasks\tabularnewline \hline
% 3.3, 4.2, 5.2: 
compositionality & -- & + & --  & analogy, abstraction, concept induction \tabularnewline 
% 3.3: 
handling data complexity & + & -- & -- & multimodal datasets\tabularnewline
\hline
% 5.3: 
memorisation of training samples & -- & -- & + & factuality datasets, precedents \tabularnewline
% 3.1, 5.3: 
explicit knowledge manipulation & -- & + & -- & causality, argumentation, theorem proving \tabularnewline
% 4.2, 5.3: 
explainable predictions & -- &  + & -- & human studies, faithfulness\tabularnewline
\bottomrule
\end{tabular*}
\end{table}

\section{Emerging Directions}
\label{sec:conclusion}

%The prior sections highlight the challenges in aligning human and machine intelligence, highlighting AI's potential to augment human generalisation capabilities. A summary of the different properties, methods, and evaluation practices for generalisation in humans and AI is shown in Table \ref{tab:summary}. The table shows the complementarity of statistical, case-to-case, and analytical generalisation approaches to satisfying desirable properties such as accuracy, shift robustness, and compositionality. The evaluation column indicates that developing adequate evaluation procedures remains challenging, especially for explainability, compositionality, and learning from a few samples. We discuss research directions toward novel theories, methods, and evaluation practices in the following.

The previous sections address the challenges of aligning human and machine intelligence, emphasising AI's potential to enhance human generalisation. Table \ref{tab:summary} summarises the properties, methods, and evaluation practices for generalisation in humans and various families of AI methods. The table highlights the complementarity of the three methods in achieving important properties for aligning human and AI generalisation. Statistical approaches enable universality of approximation and inference correctness, instance-based methods enable robustness and memorisation, while analytical techniques are designed for compositionality and explainable predictions. The evaluation column indicates the wide variety of approaches used to assess the ability of various methods, including neurosymbolic ones that aim to combine benefits across method families. The evaluation column also shows challenges in evaluating the explainability of predictions and the ability to learn from a few samples. 
% complement each other to attain , shift robustness, and compositionality. The evaluation column highlights ongoing challenges, particularly in explainability, compositionality, and learning from a few samples. 
Next, we discuss future research directions for novel generalisation theories, hybrid methods, evaluation practices, and alignment in future human-AI teams.

\paragraph{Generalisation theory in the era of foundation models}

Recent zero-shot and in-context learning approaches in LLMs implicitly generalise to tasks unrelated to their training, 
% try to generalize to tasks unrelated to their training, assuming implicit generalization to new tasks 
without explicit similarity \citep{kojima2022large,bubeck2023sparks}.\footnote{The emergence of generalisation is empirically observed more strongly in the recent large reasoning models (LRMs). LRMs extend LLMs towards producing intermediate tokens based on test-time inference scaling and post-training methods~\citep{kambhampati2025reasoning}.} In other words, model builders assume that LLMs have implicitly generalised (process, Section~\ref{sec:types_process}) to generalisations (product, Section~\ref{sec:types_product}) that allow generalisation (operator, Section~\ref{sec:types_data}) to entirely new tasks and domains. However, this assumption remains unsubstantiated, leading to an overestimation of the generalisability of foundation models and a diluting generalisability to align with empirical observations~\citep{kambhampati2025reasoning}. 
% (Section~\ref{sec:types}),
% \todo{FvH: I cannot find the argument in Section~\ref{sec:types} that backs this up?}
These observations highlight the need for further research. First, new generalisation \textit{processes and products} are needed to provide guarantees or reasons to believe that (zero-shot) application to new tasks is viable, potentially through encoding invariances/equivariances~\citep{pmlr-v48-cohenc16},
which is used in complex architectures such as AlphaFold; or through cognitively inspired representations, such as prototypes, which have proven efficient for domain generalisation~\citep{10682463}. Second, a new \textit{theory} is required to define when few- or zero-shot applications are feasible, which will likely be needed to integrate notions of ML theory with invariances or analogies between domains. It has recently been shown that \textemdash unlike in classical learning theory \textemdash high dimensionality of the signals might be key to the generalisability of few-shot learners and over-parametrised deep networks
\citep{9534395,schaeffer2024double}.

\paragraph{Generalisable neurosymbolic methods}
%Neuro-symbolic AI carries great promises, as it can combine aspects of statistical methods and analytic models, thus leading to a possible combination of statistically robust and data-driven models for complex sub-symbolic parts and explicit compositional modelling for overarching schemes.  Yet many challenges remain. First, defining \textit{provable generalisation properties}, including worst-case bounds, is essential. How can we derive formal properties for (neuro-symbolic) AI generalisation based only on compositionality rather than (weaker) statistical guarantees? Can we identify reasonable and relevant situations with formal guarantees for continual learning and learning under drift? Which surrogate cost functions are provably compatible with the underlying learning objective? Second, how to handle \textit{context} remains a question. What measures the distance between contexts? What is a reasonable projection operator for applying a generalisation from one context to another? How do we know that a context is too novel to support generalisation? When do we overgeneralise? Third, neuro-symbolic methods should \textit{represent} information efficiently and facilitate \textit{compositionality}. How can we choose a suitable representation and similarity measure(s) to enable generalisation? How do we combine multiple representations for generalisation? How do we compose two generalisations into a new one? Can we exploit compositional embeddings for compositional generalisation?
Neurosymbolic AI promises to combine statistical and analytic models, enabling robust, data-driven models for sub-symbolic parts and explicit compositional modelling for overarching schemes~\citep{DBLP:series/faia/369}. However, several challenges remain: Defining \textit{provable generalisation properties}, including worst-case bounds, is crucial. Recent work is exploiting the compositionality of neurosymbolic systems to derive formal properties for neurosymbolic generalisation 
% \todo{\url{https://arxiv.org/abs/2502.03274}}
yielding upper and lower bounds for the robustness of generalisations against disturbances in the input signal \citep{manginas2025scalable}. Taking this analysis towards verifying correctness instead of just robustness remains an unaddressed problem. 

Current symbolic representations in neurosymbolic systems are typically of low \textit{expressivity} (knowledge graphs, propositional logic), allowing for only limited forms of generalisation. Recent works have begun to explore the use of richer formalisms.  The use of description logics in neurosymbolic systems is particularly relevant for generalisation since description logics are specifically designed to capture various forms of generalisation~\citep{singhbenchmarking}. 

While a theory about the \textit{compositionality} of neurosymbolic systems is beginning to emerge~\citep{van2021modular,dedesign}, a theory on how to compose the generalisations themselves is lacking. For symbolic representations, the classical theory on abstractions~\citep{10.5555/1216075.1216082} may be a good starting point, but the question of composing latent representations, such as embeddings, remains open. 

% \textbf{TODO}: 
Finally, handling the \textit{context dependency} of generalisations remains a challenge. How do we measure the distance between contexts and apply generalisation across contexts? How do we know that a context is too novel and that we overgeneralise? A possible direction is formal modelling of contextual dimensions such as time and space, following prior research on axiomatising common-sense knowledge~\citep{lenat1998dimensions,gordon2017formal}. Here, CYC's notion of hierarchical microtheories, each containing a collection of axioms~\citep{lenat1995cyc}, can be revisited from a neurosymbolic perspective, for instance, by providing a formal hierarchy of microtheories with axioms in flexible, natural language representations, as proposed by \cite{weir2024models}.

\paragraph{Generalisation in continual learning}
The increasing availability of pre-trained models, coupled with the enormous effort required to train foundational models, has led to a transition
from homogeneous AI models, which are trained from scratch, to heterogeneous AI systems that incorporate foundation models and undergo continuous
adaptation to new data or tasks. However, naive approaches carry a high risk of
\emph{catastrophic forgetting}, which, surprisingly, seems to be higher for larger LLM models
\citep{luo2025empiricalstudycatastrophicforgetting}.

Such \emph{concept drift} can be targeted 
by  an extension of statistical methods to
hybrid approaches, which attend to the preservation or memorisation of learned signals: for example, formalising domain rules as ontologies or as symbolic constraints enables a system to detect drift whenever incoming data or model outputs violate these constraints, serving as an early warning signal for distributional change that may disrupt the generalisation capabilities of the system. Recent work on graph streams~\citep{malialis2024incremental} employs neurosymbolic prototypes, where representative subgraphs (symbolic structures) are embedded in vector spaces. 
%A loss-based drift detector triggers prototype recalculation when model performance degrades.
Another remedy can be based on data-driven approaches, such as (possibly self-supervised) rehearsal technologies, for a robust memorisation of important information, albeit at increased computational costs \citep{huang-etal-2024-mitigating}.  
More efficient alternatives aim for architectural solutions such as the incorporation of instance-based representations into statistical models
\citep{delafuente2025prototypeaugmentedhypernetworkscontinual}.
However, theoretical insight on the effect of important factors such as overparametrisation or task similarity on the generalisability and forgetting of a model is currently available for very simple models only
\citep{10.5555/3618408.3619277}.

\paragraph{Evaluation of generalisation in foundation models}

Several directions have emerged to address concerns about data contamination, spurious correlations, and overfitting in state-of-the-art models. Abstraction benchmarks for visual reasoning \citep{chollet2019measure}, analogy \citep{sourati2024arn}, and lateral thinking~\citep{jiang-etal-2023-brainteaser} are gaining popularity. Although crowd-sourcing can be used to create and scale benchmarks, it can also introduce cognitive and socio-political biases by annotators \citep{draws2021checklist}, which remains poorly understood. 
Another form of community benchmarking is early materials with informal evaluations of models' success and failure cases.
However, it remains unclear how to incorporate such examples in benchmarks.
Evaluation servers and public leaderboards with private test datasets prevent overfitting but lack standardisation and are costly to maintain. 
% (cf. \citet{arkoudas2023gpt}). 
 The future may involve moving to simulation environments and synthetic data generators \citep{duan2022survey}, though they often suffer from a sim-to-real gap. To address reproducibility, researchers proposed the model \citep{mitchell2019model} and data cards \citep{pushkarna2022data} to report the details of the experiment and reproducibility checklists based on a broad consensus \citep{kapoor2024reforms}, though their coverage of generalisability is limited.

\paragraph{Aligning generalisation in future human-AI teams} 
The achievement of effective human-AI teaming and the realisation of legal frameworks such as the EU AI Act \citep{10.1145/3665322,doi:10.1126/science.adn0117} require transparent \textit{collaboration} workflows, with \textit{explanations} bridging the gaps between human and AI reasoning \citep{akata2020research}. Section~\ref{sec_two} discussed that this alignment must occur at the output level. However, when misalignments occur, indicating a larger deviation between human and machine generalisations (\eg AI predicts tumour type 1 and the doctor diagnoses tumour type 3), designing mechanisms for realignment and error correction becomes critical. Such mechanisms pose stricter requirements for collaboration on the \textit{process level through concepts and relations} \citep{langley2013central}. Examples of realignment operationalisations are language games where realignment emerges from interaction, or physics-informed models that refine predictions on object permanence.
% Alignment of human and AI processes is a potential mechanism that can help with error correction and realignment.
% Processes that enable alignment of human and AI generalisation that help with error correction (mechanisms for re-alignment). 
% Emergent alignment - e.g., language games where this emerges from interaction. Another example: physics informed models like object permanence. Now only in a relatively narrow scenarios, we want to generalize them.
%     \item uncertainty and misalignment 

Successful collaboration at the process level requires AI to model situations (e.\,g., the affordances of objects) and other agents, as well as to learn and retain feedback indefinitely, with robust feedback mechanisms to align with human preferences 
\citep{10.5555/3692070.3692292}. 
A critical challenge of human-AI teaming is \textit{reconciling the fundamentally different reasoning paradigms} of humans and AI, such as human causal models and AI's deep learning associations. %Can these be unified into a common explanatory language? 
Efforts, such as concept-based explanations \citep{widmer2023towards} and those considering concepts and relationships \citep{finzel2024relation}, suggest the potential for intertranslatability into a common explanatory language.
% The technical realization and mediation of transparency and accountability are strengthened as requirements by legislations such as the EU's AI Act \citep{10.1145/3665322,doi:10.1126/science.adn0117}.
% AI faces legal challenges around data privacy, transparency, and accountability.
% Here, legislations such as the EU's AI Act needs to be accompanied by
% research about its technical realization and mediation of the risks posed by AI technologies is required

% \forfilip{In §6, add a paragraph on "Alignment in future human-AI teams" (likely merge with the current last par), including the following discussions:
% \begin{itemize}
%     \item Question: should we go from conveyance to inner shared (concept+relations) mental models?Example: why the system says tumor type 1 and the doctor says tumor type 3 - need to align the concepts/relations between the two; alignment of concepts and relations especially important when there is a larger deviation between human and machine generalisations.
%     \item Processes that enable alignment of human and AI generalisation that help with error correction (mechanisms for re-alignment). Emergent alignment - e.g., language games where this emerges from interaction. Another example: physics informed models like object permanence. Now only in a relatively narrow scenarios, we want to generalize them.
%     \item uncertainty and misalignment 
% \end{itemize}
% }

Meanwhile, meticulous evaluation frameworks should consider both objective task-related outcomes and subjective process-related experiences, as well as the long-term ramifications of the collaboration, taking into account each party's relative contributions and responsibilities. Despite the emergence of evaluation frameworks and metrics for humans that augment AI (\eg in manual data labelling), AI that helps humans (\eg conversational question answering), and balanced collaborations where both contribute equally (\eg medical decision making) \citep{braun2023new,li2024value,yang2024fair}, \textit{there is little research on evaluating the generalisation of such teams}.

\bmhead{Acknowledgements} 
The manuscript resulted from the Dagstuhl seminar 24192: Generalization by People and Machines, held in May 2024. %The seminar report is available at \url{https://doi.org/10.4230/DagRep.14.5.1}. 
We thank Ken Forbus, Piek Vossen, Dafna Shahaf, Wael Abd-Almageed, and Michael Waldmann, who provided valuable insights during the seminar.

\bibliographystyle{plainnat}
\bibliography{sn-bibliography}% common bib file

\begin{thebibliography}{177}
\providecommand{\natexlab}[1]{#1}
\providecommand{\url}[1]{\texttt{#1}}
\expandafter\ifx\csname urlstyle\endcsname\relax
  \providecommand{\doi}[1]{doi: #1}\else
  \providecommand{\doi}{doi: \begingroup \urlstyle{rm}\Url}\fi

\bibitem[Achtibat et~al.(2023)Achtibat, Dreyer, Eisenbraun, Bosse, Wiegand, Samek, and Lapuschkin]{achtibat2023attribution}
Reduan Achtibat, Maximilian Dreyer, Ilona Eisenbraun, Sebastian Bosse, Thomas Wiegand, Wojciech Samek, and Sebastian Lapuschkin.
\newblock From attribution maps to human-understandable explanations through concept relevance propagation.
\newblock \emph{Nature Machine Intelligence}, 5\penalty0 (9):\penalty0 1006--1019, 2023.

\bibitem[Adams et~al.(2022)Adams, Shawe-Taylor, and Guedj]{adams2022controlling}
Reuben Adams, John Shawe-Taylor, and Benjamin Guedj.
\newblock Controlling multiple errors simultaneously with a pac-bayes bound.
\newblock \emph{arXiv preprint arXiv:2202.05560}, 2022.

\bibitem[Aha(2013)]{aha2013lazy}
D.W. Aha.
\newblock \emph{Lazy Learning}.
\newblock Springer Netherlands, 2013.
\newblock ISBN 9789401720533.
\newblock URL \url{https://books.google.de/books?id=b1CqCAAAQBAJ}.

\bibitem[Akata~et al.(2020)]{akata2020research}
Zeynep Akata~et al.
\newblock A research agenda for hybrid intelligence: augmenting human intellect with collaborative, adaptive, responsible, and explainable artificial intelligence.
\newblock \emph{Computer}, 53\penalty0 (8):\penalty0 18--28, 2020.

\bibitem[Andreas(2019)]{Andreas2019MeasuringCI}
Jacob Andreas.
\newblock Measuring compositionality in representation learning.
\newblock \emph{ArXiv}, abs/1902.07181, 2019.
\newblock URL \url{https://api.semanticscholar.org/CorpusID:67749672}.

\bibitem[Atkinson et~al.(2017)Atkinson, Baroni, Giacomin, Hunter, Prakken, Reed, Simari, Thimm, and Villata]{atkinson2017towards}
Katie Atkinson, Pietro Baroni, Massimiliano Giacomin, Anthony Hunter, Henry Prakken, Chris Reed, Guillermo Simari, Matthias Thimm, and Serena Villata.
\newblock Towards artificial argumentation.
\newblock \emph{AI magazine}, 38\penalty0 (3):\penalty0 25--36, 2017.

\bibitem[Baker et~al.(2018)Baker, Pe{\~n}a, Jayamohan, and J{\'e}rusalem]{Baker2018-la}
Ruth~E Baker, Jose-Maria Pe{\~n}a, Jayaratnam Jayamohan, and Antoine J{\'e}rusalem.
\newblock Mechanistic models versus machine learning, a fight worth fighting for the biological community?
\newblock \emph{Biol Lett}, 14\penalty0 (5), May 2018.

\bibitem[Bellog\'{\i}n et~al.(2024)Bellog\'{\i}n, Grau, Larsson, Schimpf, Sengupta, and Solmaz]{10.1145/3665322}
Alejandro Bellog\'{\i}n, Oliver Grau, Stefan Larsson, Gerhard Schimpf, Biswa Sengupta, and G\"{u}rkan Solmaz.
\newblock The eu ai act and the wager on trustworthy ai.
\newblock \emph{Commun. ACM}, 67\penalty0 (12):\penalty0 58–65, November 2024.
\newblock ISSN 0001-0782.
\newblock \doi{10.1145/3665322}.
\newblock URL \url{https://doi.org/10.1145/3665322}.

\bibitem[Bengesi et~al.(2024)Bengesi, El-Sayed, Sarker, Houkpati, Irungu, and Oladunni]{bengesi2024advancements}
Staphord Bengesi, Hoda El-Sayed, Md~Kamruzzaman Sarker, Yao Houkpati, John Irungu, and Timothy Oladunni.
\newblock Advancements in generative ai: A comprehensive review of gans, gpt, autoencoders, diffusion model, and transformers.
\newblock \emph{IEEE Access}, 2024.

\bibitem[Bengio et~al.(2024)Bengio, Hinton, Yao, Song, Abbeel, Darrell, Harari, Zhang, Xue, Shalev-Shwartz, Hadfield, Clune, Maharaj, Hutter, Baydin, McIlraith, Gao, Acharya, Krueger, Dragan, Torr, Russell, Kahneman, Brauner, and Mindermann]{doi:10.1126/science.adn0117}
Yoshua Bengio, Geoffrey Hinton, Andrew Yao, Dawn Song, Pieter Abbeel, Trevor Darrell, Yuval~Noah Harari, Ya-Qin Zhang, Lan Xue, Shai Shalev-Shwartz, Gillian Hadfield, Jeff Clune, Tegan Maharaj, Frank Hutter, Atılım~Güneş Baydin, Sheila McIlraith, Qiqi Gao, Ashwin Acharya, David Krueger, Anca Dragan, Philip Torr, Stuart Russell, Daniel Kahneman, Jan Brauner, and Sören Mindermann.
\newblock Managing extreme ai risks amid rapid progress.
\newblock \emph{Science}, 384\penalty0 (6698):\penalty0 842--845, 2024.
\newblock \doi{10.1126/science.adn0117}.
\newblock URL \url{https://www.science.org/doi/abs/10.1126/science.adn0117}.

\bibitem[Bengio et~al.(2025)Bengio, Mindermann, Privitera, Besiroglu, Bommasani, Casper, Choi, Fox, Garfinkel, Goldfarb, et~al.]{bengio2025international}
Yoshua Bengio, S{\"o}ren Mindermann, Daniel Privitera, Tamay Besiroglu, Rishi Bommasani, Stephen Casper, Yejin Choi, Philip Fox, Ben Garfinkel, Danielle Goldfarb, et~al.
\newblock International ai safety report.
\newblock \emph{arXiv preprint arXiv:2501.17805}, 2025.

\bibitem[Berthet et~al.(2025)Berthet, Surbeck, and Townsend]{doi:10.1126/science.adv1170}
M.~Berthet, M.~Surbeck, and S.~W. Townsend.
\newblock Extensive compositionality in the vocal system of bonobos.
\newblock \emph{Science}, 388\penalty0 (6742):\penalty0 104--108, 2025.
\newblock \doi{10.1126/science.adv1170}.
\newblock URL \url{https://www.science.org/doi/abs/10.1126/science.adv1170}.

\bibitem[Bertsimas and Dunn(2017)]{bertsimas2017optimal}
Dimitris Bertsimas and Jack Dunn.
\newblock Optimal classification trees.
\newblock \emph{Machine Learning}, 106:\penalty0 1039--1082, 2017.

\bibitem[Biehl(2023)]{biehl2023shallow}
Michael Biehl.
\newblock \emph{The Shallow and the Deep: A biased introduction to neural networks and old school machine learning}.
\newblock University of Groningen Press, 2023.
\newblock URL \url{https://doi.org/10.21827/648c59c1a467e}.

\bibitem[Bien and Tibshirani(2011)]{bien2011prototype}
Jacob Bien and Robert Tibshirani.
\newblock Prototype selection for interpretable classification.
\newblock \emph{The Annals of Applied Statistics}, 5\penalty0 (4):\penalty0 2403--2424, 2011.

\bibitem[Bommasani~et al.(2022)]{bommasani2022opportunitiesrisksfoundationmodels}
Rishi Bommasani~et al.
\newblock On the opportunities and risks of foundation models, 2022.
\newblock URL \url{https://arxiv.org/abs/2108.07258}.

\bibitem[Boult et~al.(2019)Boult, Cruz, Dhamija, Gunther, Henrydoss, and Scheirer]{Boult2019LearningUnknown}
Terrance~E. Boult, Steve Cruz, Akshay~R. Dhamija, Manuel Gunther, James Henrydoss, and Walter~J. Scheirer.
\newblock {Learning and the Unknown : Surveying Steps Toward Open World Recognition}.
\newblock \emph{AAAI Conference on Artificial Intelligence (AAAI)}, 2019.

\bibitem[Bousquet and Elisseeff(2002)]{journals/jmlr/BousquetE02}
Olivier Bousquet and André Elisseeff.
\newblock Stability and generalization.
\newblock \emph{J. Mach. Learn. Res.}, 2:\penalty0 499--526, 2002.
\newblock URL \url{http://dblp.uni-trier.de/db/journals/jmlr/jmlr2.html#BousquetE02}.

\bibitem[Braun et~al.(2023)Braun, Greve, and Gnewuch]{braun2023new}
Marvin Braun, Maike Greve, and Ulrich Gnewuch.
\newblock The new dream team? a review of human-ai collaboration research from a human teamwork perspective.
\newblock 2023.

\bibitem[Brown~et al(2020)]{10.5555/3495724.3495883}
Tom Brown~et al.
\newblock Language models are few-shot learners.
\newblock In \emph{Proceedings of the 34th International Conference on Neural Information Processing Systems}, NIPS '20, Red Hook, NY, USA, 2020. Curran Associates Inc.
\newblock ISBN 9781713829546.

\bibitem[Bruner et~al.(1956)Bruner, Goodnow, and Austin]{bruner1956study}
Jerome Bruner, Jacqueline~J. Goodnow, and George~A. Austin.
\newblock \emph{A Study of Thinking}.
\newblock Wiley, New York, 1956.

\bibitem[Bubeck~et al(2023)]{bubeck2023sparks}
S{\'e}bastien Bubeck~et al.
\newblock Sparks of artificial general intelligence: Early experiments with gpt-4.
\newblock \emph{arXiv preprint arXiv:2303.12712}, 2023.

\bibitem[Butlin(2021)]{butlin2021ai}
Patrick Butlin.
\newblock {AI} alignment and human reward.
\newblock In \emph{{Proceedings of the 2021 AAAI/ACM Conference on AI, Ethics, and Society}}, pages 437--445, 2021.

\bibitem[Campell and Piaget(2014)]{campell2014studies}
Robert~L Campell and Jean Piaget.
\newblock \emph{Studies in reflecting abstraction}.
\newblock Psychology Press, London, 2014.

\bibitem[Cao et~al.(2024)Cao, Fang, Meng, and Liang]{10.1145/3643806}
Jiahang Cao, Jinyuan Fang, Zaiqiao Meng, and Shangsong Liang.
\newblock Knowledge graph embedding: A survey from the perspective of representation spaces.
\newblock \emph{ACM Comput. Surv.}, 56\penalty0 (6), mar 2024.
\newblock ISSN 0360-0300.
\newblock \doi{10.1145/3643806}.
\newblock URL \url{https://doi.org/10.1145/3643806}.

\bibitem[Carroll et~al.(2024)Carroll, Foote, Siththaranjan, Russell, and Dragan]{10.5555/3692070.3692292}
Micah Carroll, Davis Foote, Anand Siththaranjan, Stuart Russell, and Anca Dragan.
\newblock Ai alignment with changing and influenceable reward functions.
\newblock In \emph{Proceedings of the 41st International Conference on Machine Learning}, ICML'24. JMLR.org, 2024.

\bibitem[Chen et~al.(2020)Chen, Cheng, Juan, Wei, and Sun]{NEURIPS2020_ca4b5656}
Hung-Jen Chen, An-Chieh Cheng, Da-Cheng Juan, Wei Wei, and Min Sun.
\newblock Mitigating forgetting in online continual learning via instance-aware parameterization.
\newblock In H.~Larochelle, M.~Ranzato, R.~Hadsell, M.F. Balcan, and H.~Lin, editors, \emph{Advances in Neural Information Processing Systems}, volume~33, pages 17466--17477. Curran Associates, Inc., 2020.
\newblock URL \url{https://proceedings.neurips.cc/paper_files/paper/2020/file/ca4b5656b7e193e6bb9064c672ac8dce-Paper.pdf}.

\bibitem[Chollet(2019)]{chollet2019measure}
Fran{\c{c}}ois Chollet.
\newblock On the measure of intelligence.
\newblock \emph{arXiv preprint arXiv:1911.01547}, 2019.

\bibitem[Cohen and Welling(2016)]{pmlr-v48-cohenc16}
Taco Cohen and Max Welling.
\newblock Group equivariant convolutional networks.
\newblock In Maria~Florina Balcan and Kilian~Q. Weinberger, editors, \emph{Proceedings of The 33rd International Conference on Machine Learning}, volume~48 of \emph{Proceedings of Machine Learning Research}, pages 2990--2999, New York, New York, USA, 20--22 Jun 2016. PMLR.
\newblock URL \url{https://proceedings.mlr.press/v48/cohenc16.html}.

\bibitem[Colunga and Smith(2003)]{colung2003emergence}
Eliana Colunga and Linda~B Smith.
\newblock The emergence of abstract ideas: Evidence from networks and babies.
\newblock \emph{Philosophical Transactions of the Royal Society of London. Series B: Biological Sciences}, 358\penalty0 (1435):\penalty0 1205--1214, 2003.

\bibitem[Cossu et~al.(2025)Cossu, Bacciu, Bernardo, Valle, Gepperth, Giannini, Hammer, and Ziffer]{Cossu2025DontDA}
Andrea Cossu, Davide Bacciu, Alessio Bernardo, Emanuele~Della Valle, Alexander Gepperth, Federico Giannini, Barbara Hammer, and Giacomo Ziffer.
\newblock Don't drift away: Advances and applications of streaming and continual learning.
\newblock \emph{ESANN 2025 proceesdings}, 2025.
\newblock URL \url{https://api.semanticscholar.org/CorpusID:277835688}.

\bibitem[Cover and Hart(1967)]{cover1967nearest}
Thomas Cover and Peter Hart.
\newblock Nearest neighbor pattern classification.
\newblock \emph{IEEE Transactions on information theory}, 13\penalty0 (1):\penalty0 21--27, 1967.

\bibitem[Cover(1965)]{cover1965geometrical}
Thomas~M Cover.
\newblock Geometrical and statistical properties of systems of linear inequalities with applications in pattern recognition.
\newblock \emph{IEEE transactions on electronic computers}, \penalty0 (3):\penalty0 326--334, 1965.

\bibitem[Dalal et~al.(2024)Dalal, Rayan, Barua, Vasserman, Sarker, and Hitzler]{DBLP:conf/nesy/DalalRBVSH24}
Abhilekha Dalal, Rushrukh Rayan, Adrita Barua, Eugene~Y. Vasserman, Md.~Kamruzzaman Sarker, and Pascal Hitzler.
\newblock On the value of labeled data and symbolic methods for hidden neuron activation analysis.
\newblock In Tarek~R. Besold, Artur d'Avila Garcez, Ernesto Jim{\'{e}}nez{-}Ruiz, Roberto Confalonieri, Pranava Madhyastha, and Benedikt Wagner, editors, \emph{Neural-Symbolic Learning and Reasoning -- 18th International Conference, NeSy 2024, Barcelona, Spain, September 9-12, 2024, Proceedings, Part {II}}, volume 14980 of \emph{Lecture Notes in Computer Science}, pages 109--131, Heidelberg, 2024. Springer.
\newblock \doi{10.1007/978-3-031-71170-1\_12}.

\bibitem[Davis(2023)]{davis2023benchmarks}
Ernest Davis.
\newblock Benchmarks for automated commonsense reasoning: A survey.
\newblock \emph{ACM Computing Surveys}, 56\penalty0 (4):\penalty0 1--41, 2023.

\bibitem[de~Boer et~al.(2025)de~Boer, Smit, van Bekkum, Meyer-Vitali, and Schmid]{dedesign}
Maaike de~Boer, Quirine Smit, Michael van Bekkum, Andr{\'e} Meyer-Vitali, and Thomas Schmid.
\newblock Design patterns for llm-based neuro-symbolic systems.
\newblock \emph{Neurosymbolic Artificial Intelligence (to appear)}, 2025.

\bibitem[De~Raedt and Kersting(2008)]{de2008probabilistic}
Luc De~Raedt and Kristian Kersting.
\newblock Probabilistic inductive logic programming.
\newblock In \emph{Probabilistic inductive logic programming: theory and applications}, pages 1--27. Springer, New York, 2008.

\bibitem[Decelle(2022)]{decelle2022introduction}
Aur{\'e}lien Decelle.
\newblock An introduction to machine learning: a perspective from statistical physics.
\newblock \emph{Physica A: Statistical Mechanics and its Applications}, page 128154, 2022.

\bibitem[Dijker and Koomen(1996)]{dijker1996stereotyping}
Anton~J Dijker and Willem Koomen.
\newblock Stereotyping and attitudinal effects under time pressure.
\newblock \emph{European Journal of Social Psychology}, 26\penalty0 (1):\penalty0 61--74, 1996.

\bibitem[Dingemanse et~al.(2015)Dingemanse, Blasi, Lupyan, Christiansen, and Monaghan]{Dingemanse2015}
Mark Dingemanse, Dami{\'a}n~E. Blasi, Gary Lupyan, Morten~H. Christiansen, and Padraic Monaghan.
\newblock Arbitrariness, iconicity, and systematicity in language.
\newblock \emph{Trends in Cognitive Sciences}, 19\penalty0 (10):\penalty0 603--615, Oct 2015.
\newblock ISSN 1364-6613.
\newblock \doi{10.1016/j.tics.2015.07.013}.
\newblock URL \url{https://doi.org/10.1016/j.tics.2015.07.013}.

\bibitem[Dodge et~al.(2021)Dodge, Sap, Marasovi{\'c}, Agnew, Ilharco, Groeneveld, Mitchell, and Gardner]{dodge-etal-2021-documenting}
Jesse Dodge, Maarten Sap, Ana Marasovi{\'c}, William Agnew, Gabriel Ilharco, Dirk Groeneveld, Margaret Mitchell, and Matt Gardner.
\newblock Documenting large webtext corpora: A case study on the colossal clean crawled corpus.
\newblock In Marie-Francine Moens, Xuanjing Huang, Lucia Specia, and Scott Wen-tau Yih, editors, \emph{Proceedings of the 2021 Conference on Empirical Methods in Natural Language Processing}, pages 1286--1305, Online and Punta Cana, Dominican Republic, November 2021. Association for Computational Linguistics.
\newblock \doi{10.18653/v1/2021.emnlp-main.98}.
\newblock URL \url{https://aclanthology.org/2021.emnlp-main.98}.

\bibitem[Donnelly et~al.(2025)Donnelly, Guo, Barnett, McTavish, Chen, and Rudin]{donnelly2025rashomon}
Jon Donnelly, Zhicheng Guo, Alina~Jade Barnett, Hayden McTavish, Chaofan Chen, and Cynthia Rudin.
\newblock Rashomon sets for prototypical-part networks: Editing interpretable models in real-time.
\newblock \emph{arXiv preprint arXiv:2503.01087}, 2025.

\bibitem[Draws et~al.(2021)Draws, Rieger, Inel, Gadiraju, and Tintarev]{draws2021checklist}
Tim Draws, Alisa Rieger, Oana Inel, Ujwal Gadiraju, and Nava Tintarev.
\newblock A checklist to combat cognitive biases in crowdsourcing.
\newblock In \emph{Proceedings of the AAAI conference on human computation and crowdsourcing}, volume~9, pages 48--59, 2021.

\bibitem[Duan et~al.(2022)Duan, Yu, Tan, Zhu, and Tan]{duan2022survey}
Jiafei Duan, Samson Yu, Hui~Li Tan, Hongyuan Zhu, and Cheston Tan.
\newblock A survey of embodied ai: From simulators to research tasks.
\newblock \emph{IEEE Transactions on Emerging Topics in Computational Intelligence}, 6\penalty0 (2):\penalty0 230--244, 2022.

\bibitem[Engel and {Broeck}(2001)]{EB01}
A.~Engel and {C. van den} {Broeck}.
\newblock \emph{The Statistical Mechanics of Learning}.
\newblock Cambridge University Press, Cambridge, UK, 2001.

\bibitem[Falkenhainer et~al.(1989)Falkenhainer, Forbus, and Gentner]{falkenhainer1989structure}
Brian Falkenhainer, Kenneth~D Forbus, and Dedre Gentner.
\newblock The structure-mapping engine: Algorithm and examples.
\newblock \emph{Artificial intelligence}, 41\penalty0 (1):\penalty0 1--63, 1989.

\bibitem[Ferrara(2024)]{ferrara2024genai}
Emilio Ferrara.
\newblock Genai against humanity: Nefarious applications of generative artificial intelligence and large language models.
\newblock \emph{Journal of Computational Social Science}, pages 1--21, 2024.

\bibitem[Finzel et~al.(2024)Finzel, Hilme, Rabold, and Schmid]{finzel2024relation}
Bettina Finzel, Patrick Hilme, Johannes Rabold, and Ute Schmid.
\newblock When a relation tells more than a concept: Exploring and evaluating classifier decisions with corex.
\newblock \emph{arXiv preprint arXiv:2405.01661}, 2024.

\bibitem[Fodor and Pylyshyn(1988)]{fodor1988connectionism}
Jerry~A Fodor and Zenon~W Pylyshyn.
\newblock Connectionism and cognitive architecture: A critical analysis.
\newblock \emph{Cognition}, 28\penalty0 (1-2):\penalty0 3--71, 1988.

\bibitem[French(1995)]{french1995subtlety}
Robert~Matthew French.
\newblock \emph{The subtlety of sameness: A theory and computer model of analogy-making}.
\newblock MIT press, 1995.

\bibitem[Fuente et~al.(2025)Fuente, Pilligua, Vidal, Soutiff, Curreli, Cremers, and Barsky]{delafuente2025prototypeaugmentedhypernetworkscontinual}
Neil De~La Fuente, Maria Pilligua, Daniel Vidal, Albin Soutiff, Cecilia Curreli, Daniel Cremers, and Andrey Barsky.
\newblock Prototype augmented hypernetworks for continual learning, 2025.
\newblock URL \url{https://arxiv.org/abs/2505.07450}.

\bibitem[Gao et~al.(2024)Gao, Xiong, Gao, Jia, Pan, Bi, Dai, Sun, Wang, and Wang]{gao2024retrievalaugmentedgenerationlargelanguage}
Yunfan Gao, Yun Xiong, Xinyu Gao, Kangxiang Jia, Jinliu Pan, Yuxi Bi, Yi~Dai, Jiawei Sun, Meng Wang, and Haofen Wang.
\newblock Retrieval-augmented generation for large language models: A survey, 2024.
\newblock URL \url{https://arxiv.org/abs/2312.10997}.

\bibitem[Ge et~al.(2023)Ge, Guo, Das, Al-Garadi, and Sarker]{ge2023few}
Yao Ge, Yuting Guo, Sudeshna Das, Mohammed~Ali Al-Garadi, and Abeed Sarker.
\newblock Few-shot learning for medical text: A review of advances, trends, and opportunities.
\newblock \emph{Journal of Biomedical Informatics}, 144:\penalty0 104458, 2023.

\bibitem[Gentner(1983)]{gentner1983structure}
Dedre Gentner.
\newblock Structure-mapping: A theoretical framework for analogy.
\newblock \emph{Cognitive Science}, 7\penalty0 (2):\penalty0 155--170, 1983.

\bibitem[Gentner and Forbus(2011)]{gentner2011computational}
Dedre Gentner and Kenneth~D Forbus.
\newblock Computational models of analogy.
\newblock \emph{Wiley interdisciplinary reviews: cognitive science}, 2\penalty0 (3):\penalty0 266--276, 2011.

\bibitem[Gentner and Markman(1997)]{gentner1997structure}
Dedre Gentner and Arthur~B Markman.
\newblock Structure mapping in analogy and similarity.
\newblock \emph{American psychologist}, 52\penalty0 (1):\penalty0 45, 1997.

\bibitem[Giunchiglia et~al.(1997)Giunchiglia, Villafiorita, and Walsh]{10.5555/1216075.1216082}
Fausto Giunchiglia, Adolfo Villafiorita, and Toby Walsh.
\newblock Theories of abstraction.
\newblock \emph{AI Commun.}, 10\penalty0 (3,4):\penalty0 167–176, December 1997.
\newblock ISSN 0921-7126.

\bibitem[Gold(1967)]{GOLD1967447}
E~Mark Gold.
\newblock Language identification in the limit.
\newblock \emph{Information and Control}, 10\penalty0 (5):\penalty0 447--474, 1967.
\newblock ISSN 0019-9958.
\newblock \doi{https://doi.org/10.1016/S0019-9958(67)91165-5}.
\newblock URL \url{https://www.sciencedirect.com/science/article/pii/S0019995867911655}.

\bibitem[Gonen et~al.(2023)Gonen, Iyer, Blevins, Smith, and Zettlemoyer]{gonen-etal-2023-demystifying}
Hila Gonen, Srini Iyer, Terra Blevins, Noah Smith, and Luke Zettlemoyer.
\newblock Demystifying prompts in language models via perplexity estimation.
\newblock In Houda Bouamor, Juan Pino, and Kalika Bali, editors, \emph{Findings of the Association for Computational Linguistics: EMNLP 2023}, pages 10136--10148, Singapore, December 2023. Association for Computational Linguistics.
\newblock \doi{10.18653/v1/2023.findings-emnlp.679}.
\newblock URL \url{https://aclanthology.org/2023.findings-emnlp.679}.

\bibitem[Gordon and Hobbs(2017)]{gordon2017formal}
Andrew~S Gordon and Jerry~R Hobbs.
\newblock \emph{A formal theory of commonsense psychology: How people think people think}.
\newblock Cambridge University Press, 2017.

\bibitem[Gottweis et~al.(2025)Gottweis, Weng, Daryin, Tu, Palepu, Sirkovic, Myaskovsky, Weissenberger, Rong, Tanno, et~al.]{gottweis2025towards}
Juraj Gottweis, Wei-Hung Weng, Alexander Daryin, Tao Tu, Anil Palepu, Petar Sirkovic, Artiom Myaskovsky, Felix Weissenberger, Keran Rong, Ryutaro Tanno, et~al.
\newblock Towards an ai co-scientist.
\newblock \emph{arXiv preprint arXiv:2502.18864}, 2025.

\bibitem[Grohs and Kutyniok(2022)]{Grohs_Kutyniok_2022}
P.~Grohs and G.~Kutyniok, editors.
\newblock \emph{Mathematical Aspects of Deep Learning}.
\newblock Cambridge University Press, 2022.

\bibitem[Guha et~al.(2023)Guha, Nyarko, Ho, R{\'e}, Chilton, Chohlas-Wood, Peters, Waldon, Rockmore, Zambrano, et~al.]{guha2023legalbench}
Neel Guha, Julian Nyarko, Daniel Ho, Christopher R{\'e}, Adam Chilton, Alex Chohlas-Wood, Austin Peters, Brandon Waldon, Daniel Rockmore, Diego Zambrano, et~al.
\newblock Legalbench: A collaboratively built benchmark for measuring legal reasoning in large language models.
\newblock \emph{Advances in Neural Information Processing Systems}, 36:\penalty0 44123--44279, 2023.

\bibitem[Gulwani et~al.(2015)Gulwani, Hern{\'a}ndez-Orallo, Kitzelmann, Muggleton, Schmid, and Zorn]{gulwani2015inductive}
Sumit Gulwani, Jos{\'e} Hern{\'a}ndez-Orallo, Emanuel Kitzelmann, Stephen~H Muggleton, Ute Schmid, and Benjamin Zorn.
\newblock Inductive programming meets the real world.
\newblock \emph{Communications of the ACM}, 58\penalty0 (11):\penalty0 90--99, 2015.

\bibitem[Hammer(2000)]{10.5555/518982}
Barbara Hammer.
\newblock \emph{Learning with Recurrent Neural Networks}.
\newblock Springer-Verlag, Berlin, Heidelberg, 2000.
\newblock ISBN 185233343X.

\bibitem[Harnad(2017)]{harnad2017cognize}
S.~Harnad.
\newblock To cognize is to categorize: Cognition is categorization.
\newblock In H.~Cohen and C.~Lefebvre, editors, \emph{Handbook of Categorization in Cognitive Science (2nd edition)}, pages 21--54. Elsevier Academic Press, Amsterdam, 2017.

\bibitem[He et~al.(2022)He, Erickson, Brown, Raghunathan, and Dragan]{he2022learning}
Jerry Zhi-Yang He, Zackory Erickson, Daniel~S. Brown, Aditi Raghunathan, and Anca Dragan.
\newblock Learning representations that enable generalization in assistive tasks.
\newblock In \emph{6th Annual Conference on Robot Learning}, 2022.
\newblock URL \url{https://openreview.net/forum?id=b88HF4vd_ej}.

\bibitem[Hendrycks and Dietterich(2019)]{Hendrycks2019BenchmarkingRobustness}
Dan Hendrycks and Thomas Dietterich.
\newblock {Benchmarking neural network robustness to common corruptions and perturbations}.
\newblock \emph{International Conference on Learning Representations (ICLR)}, 2019.

\bibitem[Hitzler et~al.(2023)Hitzler, Sarker, and Eberhart]{DBLP:series/faia/369}
Pascal Hitzler, Md.~Kamruzzaman Sarker, and Aaron Eberhart, editors.
\newblock \emph{Compendium of Neurosymbolic Artificial Intelligence}, volume 369 of \emph{Frontiers in Artificial Intelligence and Applications}.
\newblock {IOS} Press, Amsterdam, 2023.
\newblock ISBN 978-1-64368-406-2.
\newblock \doi{10.3233/FAIA369}.

\bibitem[Holzinger et~al.(2023)Holzinger, Saranti, Angerschmid, Finzel, Schmid, and Mueller]{holzinger2023toward}
Andreas Holzinger, Anna Saranti, Alessa Angerschmid, Bettina Finzel, Ute Schmid, and Heimo Mueller.
\newblock Toward human-level concept learning: Pattern benchmarking for {AI} algorithms.
\newblock \emph{Patterns}, 4:\penalty0 100788, 2023.

\bibitem[Hovy and Spruit(2016)]{hovy-spruit-2016-social}
Dirk Hovy and Shannon~L. Spruit.
\newblock The social impact of natural language processing.
\newblock In Katrin Erk and Noah~A. Smith, editors, \emph{Proceedings of the 54th Annual Meeting of the Association for Computational Linguistics (Volume 2: Short Papers)}, pages 591--598, Berlin, Germany, August 2016. Association for Computational Linguistics.
\newblock \doi{10.18653/v1/P16-2096}.
\newblock URL \url{https://aclanthology.org/P16-2096}.

\bibitem[Huang et~al.(2024)Huang, Cui, Wang, Yang, Liao, Song, Yao, and Su]{huang-etal-2024-mitigating}
Jianheng Huang, Leyang Cui, Ante Wang, Chengyi Yang, Xinting Liao, Linfeng Song, Junfeng Yao, and Jinsong Su.
\newblock Mitigating catastrophic forgetting in large language models with self-synthesized rehearsal.
\newblock In Lun-Wei Ku, Andre Martins, and Vivek Srikumar, editors, \emph{Proceedings of the 62nd Annual Meeting of the Association for Computational Linguistics (Volume 1: Long Papers)}, pages 1416--1428, Bangkok, Thailand, August 2024. Association for Computational Linguistics.
\newblock \doi{10.18653/v1/2024.acl-long.77}.
\newblock URL \url{https://aclanthology.org/2024.acl-long.77/}.

\bibitem[Hunt et~al.(1966)Hunt, Marin, and Stone]{hunt1966experiments}
Earl~B Hunt, Janet Marin, and Philip~J Stone.
\newblock \emph{Experiments in Induction}.
\newblock Academic Press, New York, 1966.

\bibitem[Ilievski(2025)]{ilievski2025}
Filip Ilievski.
\newblock \emph{Human-Centric AI with Common Sense}.
\newblock Springer, 2025.

\bibitem[Ilievski et~al.(2024)Ilievski, Hammer, van Harmelen, Paassen, Saralajew, Schmid, Biehl, Bolognesi, Dong, Gashteovski, et~al.]{preprint}
Filip Ilievski, Barbara Hammer, Frank van Harmelen, Benjamin Paassen, Sascha Saralajew, Ute Schmid, Michael Biehl, Marianna Bolognesi, Xin~Luna Dong, Kiril Gashteovski, et~al.
\newblock Aligning generalisation between humans and machines.
\newblock \emph{arXiv preprint arXiv:2411.15626}, 2024.

\bibitem[Jackendoff(1985)]{jackendoff1985semantics}
Ray~S Jackendoff.
\newblock \emph{Semantics and cognition}, volume~8.
\newblock MIT press, 1985.

\bibitem[Ji et~al.(2023{\natexlab{a}})Ji, Qiu, Chen, Zhang, Lou, Wang, Duan, He, Zhou, Zhang, et~al.]{ji2023ai}
Jiaming Ji, Tianyi Qiu, Boyuan Chen, Borong Zhang, Hantao Lou, Kaile Wang, Yawen Duan, Zhonghao He, Jiayi Zhou, Zhaowei Zhang, et~al.
\newblock {AI} alignment: {A} comprehensive survey.
\newblock \emph{arXiv preprint arXiv:2310.19852}, 2023{\natexlab{a}}.

\bibitem[Ji et~al.(2023{\natexlab{b}})Ji, Lee, Frieske, Yu, Su, Xu, Ishii, Bang, Madotto, and Fung]{10.1145/3571730}
Ziwei Ji, Nayeon Lee, Rita Frieske, Tiezheng Yu, Dan Su, Yan Xu, Etsuko Ishii, Ye~Jin Bang, Andrea Madotto, and Pascale Fung.
\newblock Survey of hallucination in natural language generation.
\newblock \emph{ACM Comput. Surv.}, 55\penalty0 (12), mar 2023{\natexlab{b}}.
\newblock ISSN 0360-0300.
\newblock \doi{10.1145/3571730}.
\newblock URL \url{https://doi.org/10.1145/3571730}.

\bibitem[Jia and Liang(2017)]{jia-liang-2017-adversarial}
Robin Jia and Percy Liang.
\newblock Adversarial examples for evaluating reading comprehension systems.
\newblock In Martha Palmer, Rebecca Hwa, and Sebastian Riedel, editors, \emph{Proceedings of the 2017 Conference on Empirical Methods in Natural Language Processing}, pages 2021--2031, Copenhagen, Denmark, September 2017. Association for Computational Linguistics.
\newblock \doi{10.18653/v1/D17-1215}.
\newblock URL \url{https://aclanthology.org/D17-1215}.

\bibitem[Jiang et~al.(2023)Jiang, Ilievski, Ma, and Sourati]{jiang-etal-2023-brainteaser}
Yifan Jiang, Filip Ilievski, Kaixin Ma, and Zhivar Sourati.
\newblock {BRAINTEASER}: Lateral thinking puzzles for large language models.
\newblock In Houda Bouamor, Juan Pino, and Kalika Bali, editors, \emph{Proceedings of the 2023 Conference on Empirical Methods in Natural Language Processing}, pages 14317--14332, Singapore, December 2023. Association for Computational Linguistics.
\newblock \doi{10.18653/v1/2023.emnlp-main.885}.
\newblock URL \url{https://aclanthology.org/2023.emnlp-main.885/}.

\bibitem[Jiao et~al.(2024)Jiao, Guo, Feng, Chen, and Song]{JIAO20241}
Tianzhe Jiao, Chaopeng Guo, Xiaoyue Feng, Yuming Chen, and Jie Song.
\newblock A comprehensive survey on deep learning multi-modal fusion: Methods, technologies and applications.
\newblock \emph{Computers, Materials and Continua}, 80\penalty0 (1):\penalty0 1--35, 2024.
\newblock ISSN 1546-2218.
\newblock \doi{https://doi.org/10.32604/cmc.2024.053204}.
\newblock URL \url{https://www.sciencedirect.com/science/article/pii/S1546221824005216}.

\bibitem[Jumper~et al(2021)]{jumper2021highly}
John Jumper~et al.
\newblock Highly accurate protein structure prediction with alphafold.
\newblock \emph{nature}, 596\penalty0 (7873):\penalty0 583--589, 2021.

\bibitem[Kahneman(2011)]{kahneman2011thinking}
Daniel Kahneman.
\newblock \emph{Thinking, fast and slow}.
\newblock Macmillan, New York, 2011.

\bibitem[Kambhampati et~al.(2025)Kambhampati, Stechly, and Valmeekam]{kambhampati2025reasoning}
Subbarao Kambhampati, Kaya Stechly, and Karthik Valmeekam.
\newblock (how) do reasoning models reason?
\newblock \emph{Annals of the New York Academy of Sciences}, 2025.

\bibitem[Kapoor~et al(2024)]{kapoor2024reforms}
Sayash Kapoor~et al.
\newblock Reforms: Consensus-based recommendations for machine-learning-based science.
\newblock \emph{Science Advances}, 10\penalty0 (18):\penalty0 eadk3452, 2024.

\bibitem[Khandelwal et~al.(2020)Khandelwal, Levy, Jurafsky, Zettlemoyer, and Lewis]{Khandelwal2020Generalization}
Urvashi Khandelwal, Omer Levy, Dan Jurafsky, Luke Zettlemoyer, and Mike Lewis.
\newblock Generalization through memorization: Nearest neighbor language models.
\newblock In \emph{International Conference on Learning Representations}, 2020.
\newblock URL \url{https://openreview.net/forum?id=HklBjCEKvH}.

\bibitem[Kitzelmann and Schmid(2006)]{10.5555/1248547.1248562}
Emanuel Kitzelmann and Ute Schmid.
\newblock Inductive synthesis of functional programs: An explanation based generalization approach.
\newblock \emph{J. Mach. Learn. Res.}, 7:\penalty0 429–454, dec 2006.
\newblock ISSN 1532-4435.

\bibitem[Kojima et~al.(2022)Kojima, Gu, Reid, Matsuo, and Iwasawa]{kojima2022large}
Takeshi Kojima, Shixiang~Shane Gu, Machel Reid, Yutaka Matsuo, and Yusuke Iwasawa.
\newblock Large language models are zero-shot reasoners.
\newblock \emph{Advances in neural information processing systems}, 35:\penalty0 22199--22213, 2022.

\bibitem[Koller and Friedman(2009)]{koller2009probabilistic}
D.~Koller and N.~Friedman.
\newblock \emph{Probabilistic Graphical Models: Principles and Techniques}.
\newblock Adaptive computation and machine learning. MIT Press, 2009.
\newblock ISBN 9780262013192.
\newblock URL \url{https://books.google.co.in/books?id=7dzpHCHzNQ4C}.

\bibitem[Kulis(2013)]{MAL-019}
Brian Kulis.
\newblock Metric learning: A survey.
\newblock \emph{Foundations and Trends® in Machine Learning}, 5\penalty0 (4):\penalty0 287--364, 2013.
\newblock ISSN 1935-8237.
\newblock \doi{10.1561/2200000019}.
\newblock URL \url{http://dx.doi.org/10.1561/2200000019}.

\bibitem[Labov(1973)]{labov1973boundaries}
William Labov.
\newblock The boundaries of words and their meanings.
\newblock In C.-J.~N. Bailey and R.~W. Shuy, editors, \emph{New Ways of Analyzing Variation in English}, pages 67--90. Georgetown University Press, Washington, 1973.

\bibitem[Lafond et~al.(2009)Lafond, Lacouture, and Cohen]{lafond2009decision}
Daniel Lafond, Yves Lacouture, and Andrew~L Cohen.
\newblock Decision-tree models of categorization response times, choice proportions, and typicality judgments.
\newblock \emph{Psychological Review}, 116\penalty0 (4):\penalty0 833, 2009.

\bibitem[Lake and Baroni(2023)]{Lake2023}
Brenden~M. Lake and Marco Baroni.
\newblock Human-like systematic generalization through a meta-learning neural network.
\newblock \emph{Nature}, 623\penalty0 (7985):\penalty0 115--121, Nov 2023.
\newblock ISSN 1476-4687.
\newblock \doi{10.1038/s41586-023-06668-3}.
\newblock URL \url{https://doi.org/10.1038/s41586-023-06668-3}.

\bibitem[Lake et~al.(2015)Lake, Salakhutdinov, and Tenenbaum]{lake2015human}
Brenden~M Lake, Ruslan Salakhutdinov, and Joshua~B Tenenbaum.
\newblock Human-level concept learning through probabilistic program induction.
\newblock \emph{Science}, 350\penalty0 (6266):\penalty0 1332--1338, 2015.

\bibitem[Langley and Simon(2013)]{langley2013central}
Pat Langley and Herbert~A Simon.
\newblock The central role of learning in cognition.
\newblock In \emph{Cognitive Skills and Their Acquisition}, pages 361--380. Psychology Press, 2013.

\bibitem[Lapuschkin et~al.(2019)Lapuschkin, W{\"{a}}ldchen, Binder, Montavon, Samek, and M{\"{u}}ller]{Lapuschkin2019CleverHans}
Sebastian Lapuschkin, Stephan W{\"{a}}ldchen, Alexander Binder, Gr{\'{e}}goire Montavon, Wojciech Samek, and Klaus{-}Robert M{\"{u}}ller.
\newblock Unmasking clever hans predictors and assessing what machines really learn.
\newblock \emph{Nature Communications}, 10, 2019.

\bibitem[Lenat(1998)]{lenat1998dimensions}
Doug Lenat.
\newblock The dimensions of context-space, 1998.

\bibitem[Lenat(1995)]{lenat1995cyc}
Douglas~B Lenat.
\newblock Cyc: A large-scale investment in knowledge infrastructure.
\newblock \emph{Communications of the ACM}, 38\penalty0 (11):\penalty0 33--38, 1995.

\bibitem[Lewis and Mitchell(2024)]{lewis2024using}
Martha Lewis and Melanie Mitchell.
\newblock Using counterfactual tasks to evaluate the generality of analogical reasoning in large language models.
\newblock \emph{arXiv preprint arXiv:2402.08955}, 2024.

\bibitem[Li and Flanigan(2024)]{li2024task}
Changmao Li and Jeffrey Flanigan.
\newblock Task contamination: Language models may not be few-shot anymore.
\newblock In \emph{Proceedings of the AAAI Conference on Artificial Intelligence}, volume~38, pages 18471--18480, 2024.

\bibitem[Li et~al.(2024)Li, Liang, Peng, and Yin]{li2024value}
Zhuoyan Li, Chen Liang, Jing Peng, and Ming Yin.
\newblock The value, benefits, and concerns of generative ai-powered assistance in writing.
\newblock In \emph{Proceedings of the CHI Conference on Human Factors in Computing Systems}, pages 1--25, 2024.

\bibitem[Liao et~al.(2024)Liao, Tian, Zhang, Hua, Zou, and Li]{10682463}
Muxin Liao, Shishun Tian, Yuhang Zhang, Guoguang Hua, Wenbin Zou, and Xia Li.
\newblock Calibration-based multi-prototype contrastive learning for domain generalization semantic segmentation in traffic scenes.
\newblock \emph{IEEE Transactions on Intelligent Transportation Systems}, 25\penalty0 (12):\penalty0 20985--21001, 2024.
\newblock \doi{10.1109/TITS.2024.3454274}.

\bibitem[Lin et~al.(2017)Lin, Tegmark, and Rolnick]{Lin2017}
Henry~W. Lin, Max Tegmark, and David Rolnick.
\newblock Why does deep and cheap learning work so well?
\newblock \emph{Journal of Statistical Physics}, 168\penalty0 (6):\penalty0 1223--1247, Sep 2017.
\newblock ISSN 1572-9613.
\newblock \doi{10.1007/s10955-017-1836-5}.
\newblock URL \url{https://doi.org/10.1007/s10955-017-1836-5}.

\bibitem[Lin et~al.(2023)Lin, Ju, Liang, and Shroff]{10.5555/3618408.3619277}
Sen Lin, Peizhong Ju, Yingbin Liang, and Ness Shroff.
\newblock Theory on forgetting and generalization of continual learning.
\newblock In \emph{Proceedings of the 40th International Conference on Machine Learning}, ICML'23. JMLR.org, 2023.

\bibitem[Liu et~al.(2021)Liu, Shen, He, Zhang, Xu, Yu, and Cui]{liu2021towards}
Jiashuo Liu, Zheyan Shen, Yue He, Xingxuan Zhang, Renzhe Xu, Han Yu, and Peng Cui.
\newblock Towards out-of-distribution generalization: A survey.
\newblock \emph{arXiv preprint arXiv:2108.13624}, 2021.

\bibitem[Losing et~al.(2016)Losing, Hammer, and Wersing]{7837853}
Viktor Losing, Barbara Hammer, and Heiko Wersing.
\newblock Knn classifier with self adjusting memory for heterogeneous concept drift.
\newblock In \emph{2016 IEEE 16th International Conference on Data Mining (ICDM)}, pages 291--300, 2016.
\newblock \doi{10.1109/ICDM.2016.0040}.

\bibitem[Lu et~al.(2018)Lu, Liu, Dong, Gu, Gama, and Zhang]{lu2018learning}
Jie Lu, Anjin Liu, Fan Dong, Feng Gu, Joao Gama, and Guangquan Zhang.
\newblock Learning under concept drift: A review.
\newblock \emph{IEEE transactions on knowledge and data engineering}, 31\penalty0 (12):\penalty0 2346--2363, 2018.

\bibitem[Luo et~al.(2025)Luo, Yang, Meng, Li, Zhou, and Zhang]{luo2025empiricalstudycatastrophicforgetting}
Yun Luo, Zhen Yang, Fandong Meng, Yafu Li, Jie Zhou, and Yue Zhang.
\newblock An empirical study of catastrophic forgetting in large language models during continual fine-tuning, 2025.
\newblock URL \url{https://arxiv.org/abs/2308.08747}.

\bibitem[Malialis et~al.(2024)Malialis, Li, Panayiotou, and Polycarpou]{malialis2024incremental}
Kleanthis Malialis, Jin Li, Christos~G Panayiotou, and Marios~M Polycarpou.
\newblock Incremental learning with concept drift detection and prototype-based embeddings for graph stream classification.
\newblock In \emph{2024 International Joint Conference on Neural Networks (IJCNN)}, pages 1--7. IEEE, 2024.

\bibitem[Manginas et~al.(2025)Manginas, Manginas, Stevinson, Varghese, Katzouris, Paliouras, and Lomuscio]{manginas2025scalable}
Vasileios Manginas, Nikolaos Manginas, Edward Stevinson, Sherwin Varghese, Nikos Katzouris, Georgios Paliouras, and Alessio Lomuscio.
\newblock A scalable approach to probabilistic neuro-symbolic verification.
\newblock \emph{arXiv preprint arXiv:2502.03274}, 2025.

\bibitem[Manhaeve et~al.(2021)Manhaeve, Dumančić, Kimmig, Demeester, and {De Raedt}]{MANHAEVE2021103504}
Robin Manhaeve, Sebastijan Dumančić, Angelika Kimmig, Thomas Demeester, and Luc {De Raedt}.
\newblock Neural probabilistic logic programming in deepproblog.
\newblock \emph{Artificial Intelligence}, 298:\penalty0 103504, 2021.
\newblock ISSN 0004-3702.
\newblock \doi{https://doi.org/10.1016/j.artint.2021.103504}.
\newblock URL \url{https://www.sciencedirect.com/science/article/pii/S0004370221000552}.

\bibitem[Marcus(2003)]{marcus2003algebraic}
Gary~F Marcus.
\newblock \emph{The algebraic mind: Integrating connectionism and cognitive science}.
\newblock MIT press, 2003.

\bibitem[Medin et~al.(1987)Medin, Wattenmaker, and Hampson]{MEDIN1987242}
Douglas~L Medin, William~D Wattenmaker, and Sarah~E Hampson.
\newblock Family resemblance, conceptual cohesiveness, and category construction.
\newblock \emph{Cognitive Psychology}, 19\penalty0 (2):\penalty0 242--279, 1987.
\newblock ISSN 0010-0285.
\newblock \doi{https://doi.org/10.1016/0010-0285(87)90012-0}.
\newblock URL \url{https://www.sciencedirect.com/science/article/pii/0010028587900120}.

\bibitem[Medin et~al.(2005)Medin, Ross, and Markman]{medin2005cognitive}
{Douglas L} Medin, {Brian H} Ross, and {Arthur B} Markman.
\newblock \emph{Cognitive Psychology}.
\newblock Wiley, Hoboken, NJ, 4th edition, 2005.
\newblock ISBN 0471458201.

\bibitem[Mervis~et al(1981)]{mervis1981categorization}
Carolyn~B Mervis~et al.
\newblock Categorization of natural objects.
\newblock \emph{Annual review of psychology}, 32\penalty0 (1):\penalty0 89--115, 1981.

\bibitem[Metcalfe et~al.(2021)Metcalfe, Perelman, Boothe, and Mcdowell]{metcalfe2021systemic}
Jason~S Metcalfe, Brandon~S Perelman, David~L Boothe, and Kaleb Mcdowell.
\newblock Systemic oversimplification limits the potential for human-ai partnership.
\newblock \emph{IEEE Access}, 9:\penalty0 70242--70260, 2021.

\bibitem[Miller(2019)]{miller2019explanation}
Tim Miller.
\newblock Explanation in artificial intelligence: Insights from the social sciences.
\newblock \emph{Artificial Intelligence}, 267:\penalty0 1--38, 2019.

\bibitem[Mitchell et~al.(2019)Mitchell, Wu, Zaldivar, Barnes, Vasserman, Hutchinson, Spitzer, Raji, and Gebru]{mitchell2019model}
Margaret Mitchell, Simone Wu, Andrew Zaldivar, Parker Barnes, Lucy Vasserman, Ben Hutchinson, Elena Spitzer, Inioluwa~Deborah Raji, and Timnit Gebru.
\newblock Model cards for model reporting.
\newblock In \emph{Proceedings of the conference on fairness, accountability, and transparency}, pages 220--229, 2019.

\bibitem[Muggleton and De~Raedt(1994)]{muggleton1994inductive}
Stephen Muggleton and Luc De~Raedt.
\newblock Inductive logic programming: Theory and methods.
\newblock \emph{The Journal of Logic Programming}, 19:\penalty0 629--679, 1994.

\bibitem[M{\"u}ller et~al.(2022)M{\"u}ller, P{\'e}rez-Torr{\'o}, and Franco-Salvador]{muller-etal-2022-shot}
Thomas M{\"u}ller, Guillermo P{\'e}rez-Torr{\'o}, and Marc Franco-Salvador.
\newblock Few-shot learning with {S}iamese networks and label tuning.
\newblock In Smaranda Muresan, Preslav Nakov, and Aline Villavicencio, editors, \emph{Proceedings of the 60th Annual Meeting of the Association for Computational Linguistics (Volume 1: Long Papers)}, pages 8532--8545, Dublin, Ireland, May 2022. Association for Computational Linguistics.
\newblock \doi{10.18653/v1/2022.acl-long.584}.
\newblock URL \url{https://aclanthology.org/2022.acl-long.584/}.

\bibitem[Mundt et~al.(2023)Mundt, Hong, Pliushch, and Ramesh]{Mundt2023wholistic}
Martin Mundt, Yongwon Hong, Iuliia Pliushch, and Visvanathan Ramesh.
\newblock {A wholistic view of continual learning with deep neural networks: Forgotten lessons and the bridge to active and open world learning}.
\newblock \emph{Neural Networks}, 160:\penalty0 306--336, 2023.

\bibitem[Nauta et~al.(2023)Nauta, Trienes, Pathak, Nguyen, Peters, Schmitt, Schl\"{o}tterer, van Keulen, and Seifert]{10.1145/3583558}
Meike Nauta, Jan Trienes, Shreyasi Pathak, Elisa Nguyen, Michelle Peters, Yasmin Schmitt, J\"{o}rg Schl\"{o}tterer, Maurice van Keulen, and Christin Seifert.
\newblock From anecdotal evidence to quantitative evaluation methods: A systematic review on evaluating explainable ai.
\newblock \emph{ACM Comput. Surv.}, 55\penalty0 (13s), July 2023.
\newblock ISSN 0360-0300.
\newblock \doi{10.1145/3583558}.
\newblock URL \url{https://doi.org/10.1145/3583558}.

\bibitem[Nie et~al.(2020)Nie, Yu, Mao, Patel, Zhu, and Anandkumar]{nie2020bongard}
Weili Nie, Zhiding Yu, Lei Mao, Ankit~B Patel, Yuke Zhu, and Anima Anandkumar.
\newblock Bongard-logo: A new benchmark for human-level concept learning and reasoning.
\newblock \emph{Advances in Neural Information Processing Systems}, 33:\penalty0 16468--16480, 2020.

\bibitem[Nogueira et~al.(2022)Nogueira, Pugnana, Ruggieri, Pedreschi, and Gama]{nogueira2022methods}
Ana~Rita Nogueira, Andrea Pugnana, Salvatore Ruggieri, Dino Pedreschi, and Jo{\~a}o Gama.
\newblock Methods and tools for causal discovery and causal inference.
\newblock \emph{Wiley interdisciplinary reviews: data mining and knowledge discovery}, 12\penalty0 (2):\penalty0 e1449, 2022.

\bibitem[Nosofsky(1988)]{nosofsky1988exemplar}
Robert~M Nosofsky.
\newblock Exemplar-based accounts of relations between classification, recognition, and typicality.
\newblock \emph{Journal of Experimental Psychology: Learning, Memory, and Cognition}, 14\penalty0 (4):\penalty0 700, 1988.

\bibitem[Nosofsky(2011)]{nosofsky2011generalized}
Robert~M Nosofsky.
\newblock The generalized context model: An exemplar model of classification.
\newblock \emph{Formal approaches in categorization}, pages 18--39, 2011.

\bibitem[Papernot et~al.(2016)Papernot, McDaniel, Jha, Fredrikson, Celik, and Swami]{7467366}
Nicolas Papernot, Patrick McDaniel, Somesh Jha, Matt Fredrikson, Z.~Berkay Celik, and Ananthram Swami.
\newblock The limitations of deep learning in adversarial settings.
\newblock In \emph{2016 IEEE European Symposium on Security and Privacy (EuroS\&P)}, pages 372--387, 2016.
\newblock \doi{10.1109/EuroSP.2016.36}.

\bibitem[Peterson(2009)]{peterson2009k}
Leif~E Peterson.
\newblock K-nearest neighbor.
\newblock \emph{Scholarpedia}, 4\penalty0 (2):\penalty0 1883, 2009.

\bibitem[Pushkarna et~al.(2022)Pushkarna, Zaldivar, and Kjartansson]{pushkarna2022data}
Mahima Pushkarna, Andrew Zaldivar, and Oddur Kjartansson.
\newblock {Data Cards: Purposeful and Transparent Dataset Documentation for Responsible AI}.
\newblock In \emph{Proceedings of the 2022 ACM Conference on Fairness, Accountability, and Transparency}, pages 1776--1826, 2022.

\bibitem[Quinlan(1987)]{quinlan1987simplifying}
J.~Ross Quinlan.
\newblock Simplifying decision trees.
\newblock \emph{International Journal of Man-Machine Studies}, 27\penalty0 (3):\penalty0 221--234, 1987.

\bibitem[Reilly~et al.(2003)]{reilly2003we}
Jamie Reilly~et al.
\newblock What we mean when we say semantic: A consensus statement on the nomenclature of semantic memory.
\newblock \emph{PsyArXiv preprint: https://osf. io/pre prints/psyarxiv/xrnb2}, 2003.

\bibitem[Rosch and Mervis(1975)]{rosch1975family}
Eleanor Rosch and Carolyn~B Mervis.
\newblock Family resemblances: Studies in the internal structure of categories.
\newblock \emph{Cognitive psychology}, 7\penalty0 (4):\penalty0 573--605, 1975.

\bibitem[Rosch(1973)]{ROSCH1973328}
Eleanor~H. Rosch.
\newblock Natural categories.
\newblock \emph{Cognitive Psychology}, 4\penalty0 (3):\penalty0 328--350, 1973.
\newblock ISSN 0010-0285.
\newblock \doi{https://doi.org/10.1016/0010-0285(73)90017-0}.
\newblock URL \url{https://www.sciencedirect.com/science/article/pii/0010028573900170}.

\bibitem[Rumelhart et~al.(1986)Rumelhart, McClelland, and Group]{rumelhart1986parallel}
David~E. Rumelhart, James~L. McClelland, and PDP~Research Group.
\newblock \emph{Parallel distributed processing, volume 1: Explorations in the microstructure of cognition: Foundations}.
\newblock The MIT press, Cambridge, MA, 1986.

\bibitem[Schaeffer et~al.(2024)Schaeffer, Robertson, Boopathy, Khona, Pistunova, Rocks, Fiete, Gromov, and Koyejo]{schaeffer2024double}
Rylan Schaeffer, Zachary Robertson, Akhilan Boopathy, Mikail Khona, Kateryna Pistunova, Jason~William Rocks, Ila~R Fiete, Andrey Gromov, and Sanmi Koyejo.
\newblock Double descent demystified: Identifying, interpreting \& ablating the sources of a deep learning puzzle.
\newblock In \emph{The Third Blogpost Track at ICLR 2024}, 2024.
\newblock URL \url{https://openreview.net/forum?id=muC7uLvGHr}.

\bibitem[Schmid and Kitzelmann(2011)]{schmid2011inductive}
Ute Schmid and Emanuel Kitzelmann.
\newblock Inductive rule learning on the knowledge level.
\newblock \emph{Cognitive Systems Research}, 12\penalty0 (3-4):\penalty0 237--248, 2011.

\bibitem[Schneider et~al.(2009)Schneider, Biehl, and Hammer]{DBLP:journals/neco/SchneiderBH09a}
Petra Schneider, Michael Biehl, and Barbara Hammer.
\newblock Adaptive relevance matrices in {Learning Vector Quantization}.
\newblock \emph{Neural Comput.}, 21\penalty0 (12):\penalty0 3532--3561, 2009.
\newblock \doi{10.1162/NECO.2009.11-08-908}.
\newblock URL \url{https://doi.org/10.1162/neco.2009.11-08-908}.

\bibitem[Sch{\"o}lkopf et~al.(2021)Sch{\"o}lkopf, Locatello, Bauer, Ke, Kalchbrenner, Goyal, and Bengio]{scholkopf2021toward}
Bernhard Sch{\"o}lkopf, Francesco Locatello, Stefan Bauer, Nan~Rosemary Ke, Nal Kalchbrenner, Anirudh Goyal, and Yoshua Bengio.
\newblock Toward causal representation learning.
\newblock \emph{Proceedings of the IEEE}, 109\penalty0 (5):\penalty0 612--634, 2021.

\bibitem[Sehgal et~al.(2024)Sehgal, Grayeli, Sun, and Chaudhuri]{DBLP:conf/iclr/SehgalGSC24}
Atharva Sehgal, Arya Grayeli, Jennifer~J. Sun, and Swarat Chaudhuri.
\newblock Neurosymbolic grounding for compositional world models.
\newblock In \emph{The Twelfth International Conference on Learning Representations, {ICLR} 2024, Vienna, Austria, May 7-11, 2024}. OpenReview.net, 2024.
\newblock URL \url{https://openreview.net/forum?id=4KZpDGD4Nh}.

\bibitem[Seung et~al.(1992)Seung, Sompolinsky, and Tishby]{PhysRevA.45.6056}
H.~S. Seung, H.~Sompolinsky, and N.~Tishby.
\newblock Statistical mechanics of learning from examples.
\newblock \emph{Phys. Rev. A}, 45:\penalty0 6056--6091, Apr 1992.
\newblock \doi{10.1103/PhysRevA.45.6056}.
\newblock URL \url{https://link.aps.org/doi/10.1103/PhysRevA.45.6056}.

\bibitem[Shah et~al.(2020)Shah, Tamuly, Raghunathan, Jain, and Netrapalli]{Shay2020SimplicityBias}
Harshay Shah, Kaustav Tamuly, Aditi Raghunathan, Prateek Jain, and Praneeth Netrapalli.
\newblock The pitfalls of simplicity bias in neural networks.
\newblock \emph{Neural Information Processing Systems}, 2020.

\bibitem[Shalev-Shwartz and Ben-David(2014)]{shalev2014understanding}
Shai Shalev-Shwartz and Shai Ben-David.
\newblock \emph{Understanding machine learning: From theory to algorithms}.
\newblock Cambridge university press, 2014.

\bibitem[Shepard(1987)]{doi:10.1126/science.3629243}
Roger~N. Shepard.
\newblock Toward a universal law of generalization for psychological science.
\newblock \emph{Science}, 237\penalty0 (4820):\penalty0 1317--1323, 1987.
\newblock \doi{10.1126/science.3629243}.
\newblock URL \url{https://www.science.org/doi/abs/10.1126/science.3629243}.

\bibitem[Shin and Kaneko(2024)]{shin2024large}
Andrew Shin and Kunitake Kaneko.
\newblock Large language models lack understanding of character composition of words, 2024.

\bibitem[Singh et~al.(2025)Singh, Tommasini, Bhatia, and Mutharaju]{singhbenchmarking}
Gunjan Singh, Riccardo Tommasini, Sumit Bhatia, and Raghava Mutharaju.
\newblock Benchmarking neuro-symbolic description logic reasoners: Existing challenges and a way forward.
\newblock \emph{Neurosymbolic Artificial Intelligence (to appear)}, 2025.

\bibitem[Son et~al.(2008)Son, Smith, and Goldstone]{son2008simplicity}
Ji~Y Son, Linda~B Smith, and Robert~L Goldstone.
\newblock Simplicity and generalization: Short-cutting abstraction in children’s object categorizations.
\newblock \emph{Cognition}, 108\penalty0 (3):\penalty0 626--638, 2008.

\bibitem[Sourati et~al.(2023)Sourati, Venkatesh, Deshpande, Rawlani, Ilievski, Sandlin, and Mermoud]{sourati2023robust}
Zhivar Sourati, Vishnu Priya~Prasanna Venkatesh, Darshan Deshpande, Himanshu Rawlani, Filip Ilievski, H{\^o}ng-{\^A}n Sandlin, and Alain Mermoud.
\newblock Robust and explainable identification of logical fallacies in natural language arguments.
\newblock \emph{Knowledge-Based Systems}, 266:\penalty0 110418, 2023.

\bibitem[Sourati et~al.(2024)Sourati, Ilievski, Sommerauer, and Jiang]{sourati2024arn}
Zhivar Sourati, Filip Ilievski, Pia Sommerauer, and Yifan Jiang.
\newblock Arn: Analogical reasoning on narratives.
\newblock \emph{Transactions of the Association for Computational Linguistics}, 12:\penalty0 1063--1086, 2024.

\bibitem[Straat et~al.(2022)Straat, Abadi, Kan, G{\"o}pfert, Hammer, and Biehl]{straat2022supervised}
Michiel Straat, Fthi Abadi, Zhuoyun Kan, Christina G{\"o}pfert, Barbara Hammer, and Michael Biehl.
\newblock Supervised learning in the presence of concept drift: a modelling framework.
\newblock \emph{Neural Computing and Applications}, 34\penalty0 (1):\penalty0 101--118, 2022.

\bibitem[Tenenbaum and Griffiths(2001)]{Tenenbaum_Griffiths_2001}
Joshua~B. Tenenbaum and Thomas~L. Griffiths.
\newblock Generalization, similarity, and bayesian inference.
\newblock \emph{Behavioral and Brain Sciences}, 24\penalty0 (4):\penalty0 629–640, 2001.
\newblock \doi{10.1017/S0140525X01000061}.

\bibitem[Tenenbaum et~al.(2011)Tenenbaum, Kemp, Griffiths, and Goodman]{tenenbaum2011grow}
Joshua~B Tenenbaum, Charles Kemp, Thomas~L Griffiths, and Noah~D Goodman.
\newblock How to grow a mind: Statistics, structure, and abstraction.
\newblock \emph{Science}, 331\penalty0 (6022):\penalty0 1279--1285, 2011.

\bibitem[Tripuraneni et~al.(2020)Tripuraneni, Jordan, and Jin]{tripuraneni2020theory}
Nilesh Tripuraneni, Michael Jordan, and Chi Jin.
\newblock On the theory of transfer learning: The importance of task diversity.
\newblock \emph{Advances in neural information processing systems}, 33:\penalty0 7852--7862, 2020.

\bibitem[Tyukin et~al.(2021)Tyukin, Gorban, Alkhudaydi, and Zhou]{9534395}
Ivan~Y. Tyukin, Alexander~N. Gorban, Muhammad~H. Alkhudaydi, and Qinghua Zhou.
\newblock Demystification of few-shot and one-shot learning.
\newblock In \emph{2021 International Joint Conference on Neural Networks (IJCNN)}, pages 1--7, 2021.
\newblock \doi{10.1109/IJCNN52387.2021.9534395}.

\bibitem[Usynin et~al.(2024)Usynin, Knolle, and Kaissis]{usynin2024memorisation}
Dmitrii Usynin, Moritz Knolle, and Georgios Kaissis.
\newblock Memorisation in machine learning: A survey of results.
\newblock \emph{Transactions on Machine Learning Research}, 2024.

\bibitem[Vaccaro et~al.(2024)Vaccaro, Almaatouq, and Malone]{vaccaro2024combinations}
Michelle Vaccaro, Abdullah Almaatouq, and Thomas Malone.
\newblock When combinations of humans and ai are useful: {A} systematic review and meta-analysis.
\newblock \emph{Nature Human Behaviour}, pages 1--11, 2024.

\bibitem[van Bekkum et~al.(2021)van Bekkum, de~Boer, van Harmelen, Meyer-Vitali, and Teije]{van2021modular}
Michael van Bekkum, Maaike de~Boer, Frank van Harmelen, Andr{\'e} Meyer-Vitali, and Annette~ten Teije.
\newblock Modular design patterns for hybrid learning and reasoning systems: a taxonomy, patterns and use cases.
\newblock \emph{Applied Intelligence}, 51\penalty0 (9):\penalty0 6528--6546, 2021.

\bibitem[Vapnik(1995)]{vapnik95}
Vladimir~N. Vapnik.
\newblock \emph{The nature of statistical learning theory}.
\newblock Springer-Verlag New York, Inc., 1995.
\newblock ISBN 0-387-94559-8.

\bibitem[Vapnik and Chervonenkis(2015)]{vapnik2015uniform}
Vladimir~N Vapnik and A~Ya Chervonenkis.
\newblock On the uniform convergence of relative frequencies of events to their probabilities.
\newblock In \emph{Measures of complexity: festschrift for alexey chervonenkis}, pages 11--30. Springer, 2015.

\bibitem[Vats~et al(2024)]{vats2024survey}
Vanshika Vats~et al.
\newblock A survey on human-ai teaming with large pre-trained models.
\newblock \emph{arXiv preprint arXiv:2403.04931}, 2024.

\bibitem[Verwimp~et al(2024)]{verwimp2023continual}
Eli Verwimp~et al.
\newblock Continual learning: Applications and the road forward.
\newblock \emph{Transactions on Machine Learning Research}, 2024.
\newblock ISSN 2835-8856.
\newblock URL \url{https://openreview.net/forum?id=axBIMcGZn9}.

\bibitem[Waldmann et~al.(2006)Waldmann, Hagmayer, and Blaisdell]{waldmann2006beyond}
Michael~R Waldmann, York Hagmayer, and Aaron~P Blaisdell.
\newblock Beyond the information given: Causal models in learning and reasoning.
\newblock \emph{Current Directions in Psychological Science}, 15\penalty0 (6):\penalty0 307--311, 2006.

\bibitem[Wang et~al.(2023)Wang, Liu, Yue, Tang, Zhang, Jiayang, Yao, Gao, Hu, Qi, et~al.]{wang2023survey}
Cunxiang Wang, Xiaoze Liu, Yuanhao Yue, Xiangru Tang, Tianhang Zhang, Cheng Jiayang, Yunzhi Yao, Wenyang Gao, Xuming Hu, Zehan Qi, et~al.
\newblock Survey on factuality in large language models: Knowledge, retrieval and domain-specificity.
\newblock \emph{arXiv preprint arXiv:2310.07521}, 2023.

\bibitem[Weir et~al.(2024)Weir, Mishra, Weller, Tafjord, Hornstein, Sabol, Jansen, Van~Durme, and Clark]{weir2024models}
Nathaniel Weir, Bhavana~Dalvi Mishra, Orion Weller, Oyvind Tafjord, Sam Hornstein, Alexander Sabol, Peter Jansen, Benjamin Van~Durme, and Peter Clark.
\newblock From models to microtheories: Distilling a model's topical knowledge for grounded question answering.
\newblock \emph{arXiv preprint arXiv:2412.17701}, 2024.

\bibitem[Widmer et~al.(2023)Widmer, Sarker, Nadella, Fiechter, Juvina, Minnery, Hitzler, Schwartz, and Raymer]{widmer2023towards}
Cara~Leigh Widmer, Md~Kamruzzaman Sarker, Srikanth Nadella, Joshua Fiechter, Ion Juvina, Brandon Minnery, Pascal Hitzler, Joshua Schwartz, and Michael Raymer.
\newblock Towards human-compatible xai: Explaining data differentials with concept induction over background knowledge.
\newblock \emph{Journal of Web Semantics}, 79:\penalty0 100807, 2023.

\bibitem[Wiedemer et~al.(2024)Wiedemer, Brady, Panfilov, Juhos, Bethge, and Brendel]{wiedemer2024provable}
Thadd{\"a}us Wiedemer, Jack Brady, Alexander Panfilov, Attila Juhos, Matthias Bethge, and Wieland Brendel.
\newblock Provable compositional generalization for object-centric learning.
\newblock In \emph{The Twelfth International Conference on Learning Representations}, 2024.
\newblock URL \url{https://openreview.net/forum?id=7VPTUWkiDQ}.

\bibitem[Wiese et~al.(2008)Wiese, Konderding, and Schmid]{wiese2008mapping}
Eva Wiese, Uwe Konderding, and Ute Schmid.
\newblock Mapping and inference in analogical problem solving—as much as needed or as much as possible?
\newblock In \emph{Proceedings of the Annual Meeting of the Cognitive Science Society}, volume~30, pages 927--932, 2008.

\bibitem[Winston(1970)]{winston1970learning}
Patrick~H Winston.
\newblock Learning structural descriptions from examples.
\newblock Technical report, MIT, AI Technical Reports, 1970.

\bibitem[Wolpert and Macready(1997)]{wolpert1997no}
David~H Wolpert and William~G Macready.
\newblock No free lunch theorems for optimization.
\newblock \emph{IEEE transactions on evolutionary computation}, 1\penalty0 (1):\penalty0 67--82, 1997.

\bibitem[Yang et~al.(2024{\natexlab{a}})Yang, Zhou, Li, and Liu]{yang2024generalized}
Jingkang Yang, Kaiyang Zhou, Yixuan Li, and Ziwei Liu.
\newblock Generalized out-of-distribution detection: A survey.
\newblock \emph{International Journal of Computer Vision}, pages 1--28, 2024{\natexlab{a}}.

\bibitem[Yang et~al.(2023)Yang, Swope, Gu, Chalamala, Song, Yu, Godil, Prenger, and Anandkumar]{yang2023leandojo}
Kaiyu Yang, Aidan Swope, Alex Gu, Rahul Chalamala, Peiyang Song, Shixing Yu, Saad Godil, Ryan~J Prenger, and Animashree Anandkumar.
\newblock Leandojo: Theorem proving with retrieval-augmented language models.
\newblock \emph{Advances in Neural Information Processing Systems}, 36:\penalty0 21573--21612, 2023.

\bibitem[Yang et~al.(2024{\natexlab{b}})Yang, Arai, Yamashita, and Baba]{yang2024fair}
Mingzhe Yang, Hiromi Arai, Naomi Yamashita, and Yukino Baba.
\newblock Fair machine guidance to enhance fair decision making in biased people.
\newblock In \emph{Proceedings of the CHI Conference on Human Factors in Computing Systems}, pages 1--18, 2024{\natexlab{b}}.

\bibitem[Yao et~al.(2021)Yao, Chu, Li, Li, Gao, and Zhang]{10.1145/3444944}
Liuyi Yao, Zhixuan Chu, Sheng Li, Yaliang Li, Jing Gao, and Aidong Zhang.
\newblock A survey on causal inference.
\newblock \emph{ACM Trans. Knowl. Discov. Data}, 15\penalty0 (5), may 2021.
\newblock ISSN 1556-4681.
\newblock \doi{10.1145/3444944}.
\newblock URL \url{https://doi.org/10.1145/3444944}.

\bibitem[Ye et~al.(2021)Ye, Xie, Cai, Li, Li, and Wang]{ye2021towards}
Haotian Ye, Chuanlong Xie, Tianle Cai, Ruichen Li, Zhenguo Li, and Liwei Wang.
\newblock Towards a theoretical framework of out-of-distribution generalization.
\newblock In A.~Beygelzimer, Y.~Dauphin, P.~Liang, and J.~Wortman Vaughan, editors, \emph{Advances in Neural Information Processing Systems}, 2021.
\newblock URL \url{https://openreview.net/forum?id=kFJoj7zuDVi}.

\bibitem[Yuan et~al.(2025)Yuan, Li, and Zhao]{10.1145/3713070}
Yuan Yuan, Zhaojian Li, and Bin Zhao.
\newblock A survey of multimodal learning: Methods, applications, and future.
\newblock \emph{ACM Comput. Surv.}, 57\penalty0 (7), February 2025.
\newblock ISSN 0360-0300.
\newblock \doi{10.1145/3713070}.
\newblock URL \url{https://doi.org/10.1145/3713070}.

\bibitem[Zeugmann(2003)]{Zeugmann}
Thomas Zeugmann.
\newblock Can learning in the limit be done efficiently?
\newblock In Ricard Gavald{\'a}, Klaus~P. Jantke, and Eiji Takimoto, editors, \emph{Algorithmic Learning Theory}, pages 17--38, Berlin, Heidelberg, 2003. Springer Berlin Heidelberg.
\newblock ISBN 978-3-540-39624-6.

\bibitem[Zhang and Yang(2021)]{zhang2021survey}
Yu~Zhang and Qiang Yang.
\newblock A survey on multi-task learning.
\newblock \emph{IEEE transactions on knowledge and data engineering}, 34\penalty0 (12):\penalty0 5586--5609, 2021.

\bibitem[Zhuang et~al.(2020)Zhuang, Qi, Duan, Xi, Zhu, Zhu, Xiong, and He]{zhuang2020comprehensive}
Fuzhen Zhuang, Zhiyuan Qi, Keyu Duan, Dongbo Xi, Yongchun Zhu, Hengshu Zhu, Hui Xiong, and Qing He.
\newblock A comprehensive survey on transfer learning.
\newblock \emph{Proceedings of the IEEE}, 109\penalty0 (1):\penalty0 43--76, 2020.

\end{thebibliography}

\end{document}